\documentclass[lettersize,journal]{IEEEtran}
\usepackage{amsmath,amsfonts}
\usepackage{algorithm}
\usepackage{array}
\usepackage[caption=false,font=normalsize,labelfont=sf,textfont=sf]{subfig}
\usepackage{textcomp}
\usepackage{stfloats}
\usepackage{url}
\usepackage{verbatim}
\usepackage{graphicx}
\usepackage{cite}
\usepackage{color}
\newtheorem{corollary}{\bf Corollary}
\newtheorem{assumption}{\bf Assumption}

\usepackage{algorithmicx, algpseudocode}
\usepackage{listings}

\usepackage{bbding}
\usepackage{threeparttable}
\usepackage{booktabs}

\definecolor{dkgreen}{RGB}{98,151,85}

\begin{document}

\title{Phase Matching for Out-of-Distribution Generalization}

\author{Chengming Hu, Yeqian Du, Rui Wang, Hao Chen, and Congcong Zhu~\IEEEmembership{Member, ~IEEE} 
\thanks{Chengming Hu conceived of the presented idea, completed coding and developed the theoretical formalism. Yeqian Du, Rui Wang revised the first manuscript. Hao Chen provided insights on the theoretical formalism. Congcong Zhu approved the final manuscript and ensured the precision and appropriateness of the references. Congcong Zhu supported grants for this work. All authors discussed the results and commented on the manuscript. 

Chengming Hu, Yeqian Du, Rui Wang, Hao Chen and Congcong Zhu are with the University of Science and Technology of China, Hefei 230000, China. Congcong Zhu also is with  Suzhou Institute for Advanced Research, USTC, Suzhou 215123, China (Email:\{cmhu, duyeqian, rui\_wang, ch330822\}@mail.ustc.edu.cn; cczly@ustc.edu.cn)

(Corresponding author: Congcong Zhu.)  
}}

\markboth{Journal of \LaTeX\ Class Files,~Vol.~14, No.~8, August~2021}%
{Shell \MakeLowercase{\textit{et al.}}: A Sample Article Using IEEEtran.cls for IEEE Journals}


\maketitle
 
\begin{abstract}
The Fourier transform, an explicit decomposition method for visual signals, has been employed to explain the out-of-distribution generalization behaviors of Deep Neural Networks (DNNs). Previous studies indicate that the amplitude spectrum is susceptible to the disturbance caused by distribution shifts, whereas the phase spectrum preserves highly-structured spatial information that is crucial for robust visual representation learning. Inspired by this insight, this paper is dedicated to clarifying the relationships between Domain Generalization (DG) and the frequency components. Specifically, we provide distribution analysis and empirical experiments for the frequency components. Based on these observations, we propose a Phase Matching approach, termed PhaMa, to address DG problems. To this end, PhaMa introduces perturbations on the amplitude spectrum and establishes spatial relationships to match the phase components with patch contrastive learning. Experiments on multiple benchmarks demonstrate that our proposed method achieves state-of-the-art performance in domain generalization and out-of-distribution robustness tasks. Beyond vanilla analysis and experiments, we further clarify the relationships between the Fourier components and DG problems by introducing a Fourier-based Structural Causal Model (SCM).
\end{abstract}


\section{Introduction}
\label{sec:intro}
\IEEEPARstart{B}{y} assuming the typical independent and identically distributed (i.i.d.) setting for training and testing data\cite{lecun2015deep,goodfellow2016deep}, Deep Neural Networks (DNNs) have demonstrated exceptional performance on various visual tasks. However, in real-world scenarios, the DNNs often exhibit subpar performance due to the unknown distribution shifts, also known as domain shifts, between the training and testing data. Consequently, Domain Generalization (DG)\cite{muandet2013domain} has recently attracted increasing attention, an approach that aims to enable machine learning models to generalize to unseen data distributions.

\begin{figure}[!t]
\centering
\subfloat[$F_{c}$.]{\includegraphics[width=0.45\linewidth]{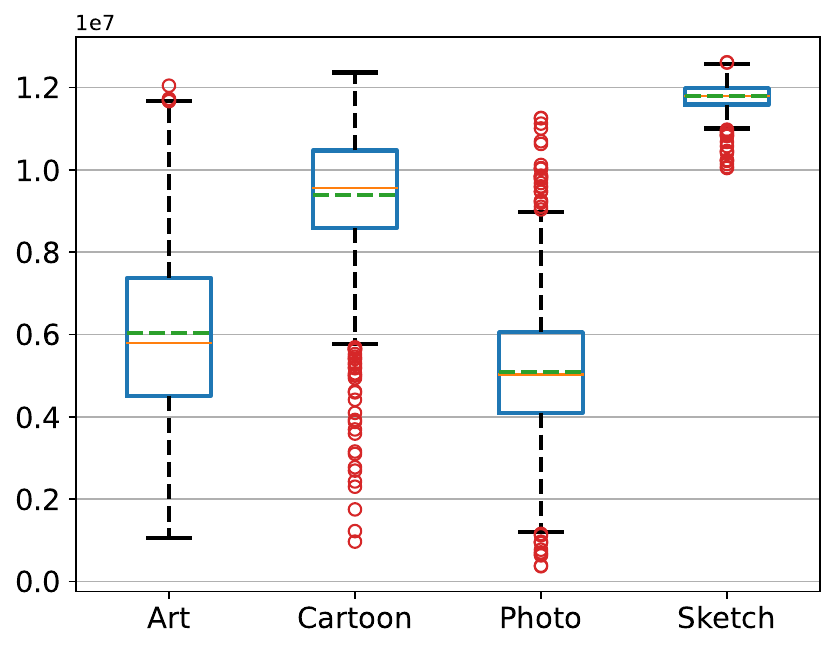}%
\label{fig:pacs_sta_fc}}
\hfil
\subfloat[$F_{std}$.]{\includegraphics[width=0.45\linewidth]{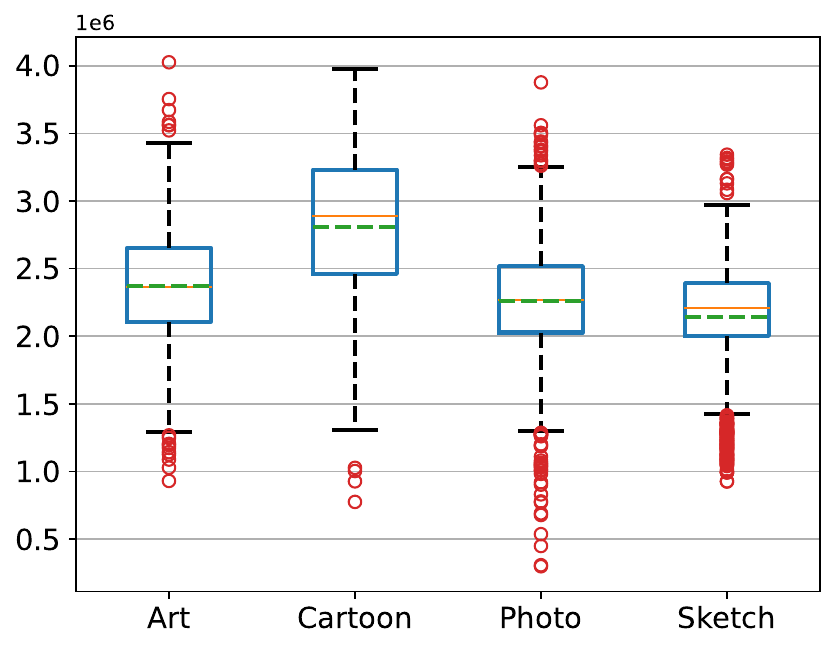}%
\label{fig:pacs_sta_fstd}}

\subfloat[PACS benchmark.]{\includegraphics[width=0.9\linewidth]{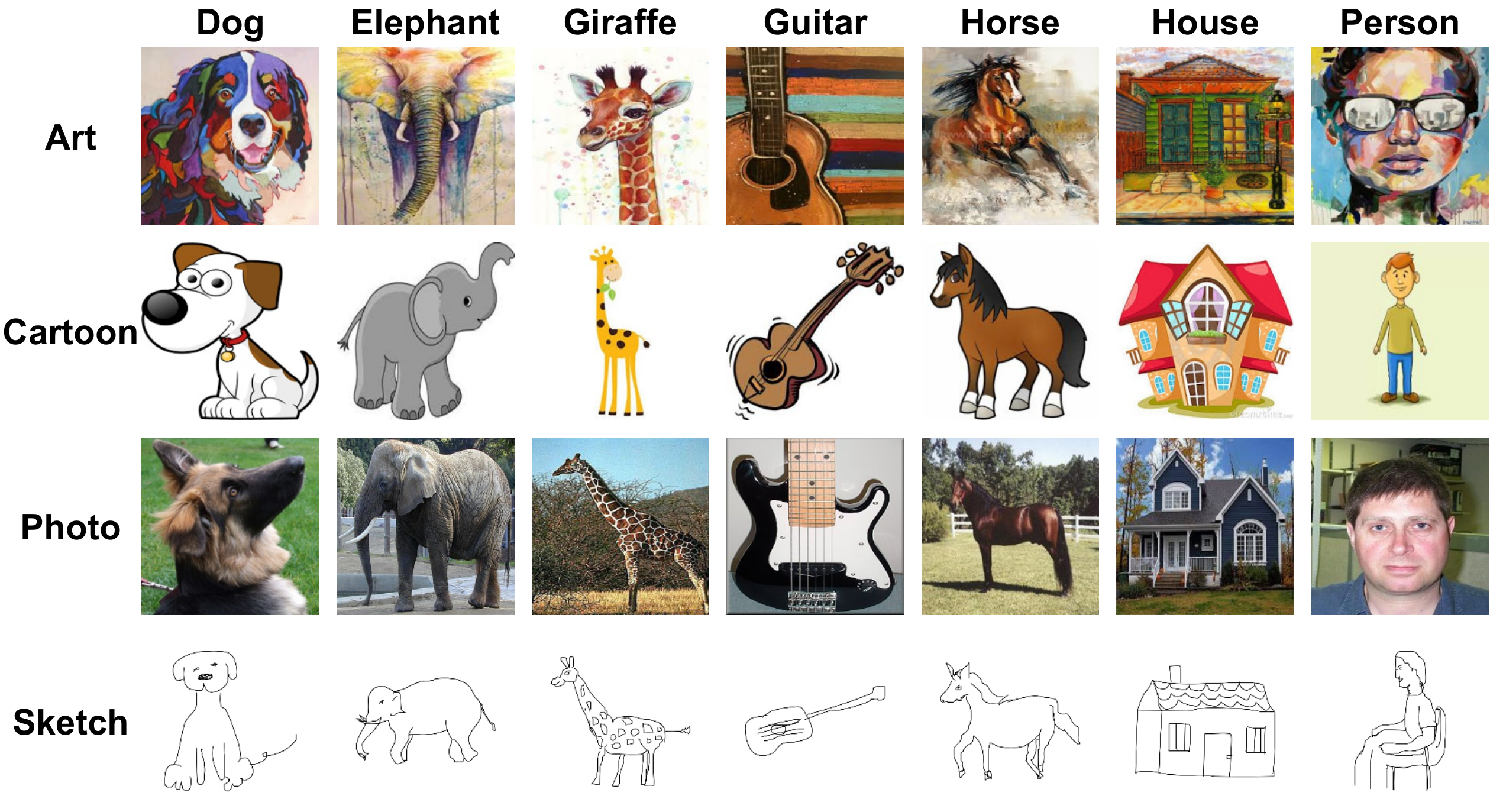}%
\label{fig:pacs}}
\caption{(a) and (b) are the boxplots of the centroid frequency $F_{c}$ and frequency standard deviation $F_{std}$ of the amplitude spectra from PACS. (c) PACS dataset\cite{li2017deeper} is a commonly used benchmark for domain generalization, comprising of four distinct domains: Art Painting, Cartoon, Photo, and Sketch.}
\label{fig:1}
\end{figure}

Mainstream Domain Generalization studies \cite{arjovsky2019invariant,li2018domain,mahajan2021domain,ilse2020diva,peng2019domain,lv2022causality} primarily focus on extracting invariant representations from source domains that can be effectively generalized to the target domain, which remains inaccessible during training. Another branch involves data augmentation\cite{hendrycks2019augmix,yun2019cutmix,zhang2018mixup,huang2017arbitrary,nuriel2021permuted,zhou2021domain,li2022uncertainty}, a technique to simulate domain shifts or attacks without changing the label. Data augmentation can also be viewed as an approach to compel the network to extract invariant representations under various perturbations (\textit{e.g.}, flip, brightness, contrast, and style).

Recent studies\cite{chen2021amplitude,guo2018low,liu2021spatial,sharma2019effectiveness,wang2020high} have focused on exploring the explanations for DNN's generalizability in the frequency domain. 
Because the Fourier transform has a well-known property: the phase spectrum preserves high-level semantics of the image while the amplitude spectrum contains low-level statistics\cite{hansen2007structural,oppenheim1979phase,oppenheim1981importance,yang2020fda}. Experiments conducted in previous studies have demonstrated the sensitivity of DNNs to the amplitude spectrum, which degrades the network's performance on out-of-distribution data. To visually explore the relationships between the amplitude spectrum and domain shifts, we calculate the centroid frequency $F_{c}$  and frequency standard deviation $F_{std}$ for the amplitude spectra, which can be simply understood as the mean and variance in the frequency domain, respectively (Detailed definition is provided in Sec. \ref{app:sta_aly}). The statistics presented in Figs. \ref{fig:pacs_sta_fc} and \ref{fig:pacs_sta_fstd} show that Art Painting and Photo show a substantial overlap. In contrast, Cartoon and Sketch exhibit significant distribution shifts. These results align with the images presented in Fig. \ref{fig:pacs}, where Cartoon and Sketch significantly differ from the others, while the color and edge variations in both Photo and Art Painting are quite complex.

\begin{figure*}[!t]
\centering
\subfloat[Amplitude Only.]{\includegraphics[width=0.49\linewidth]{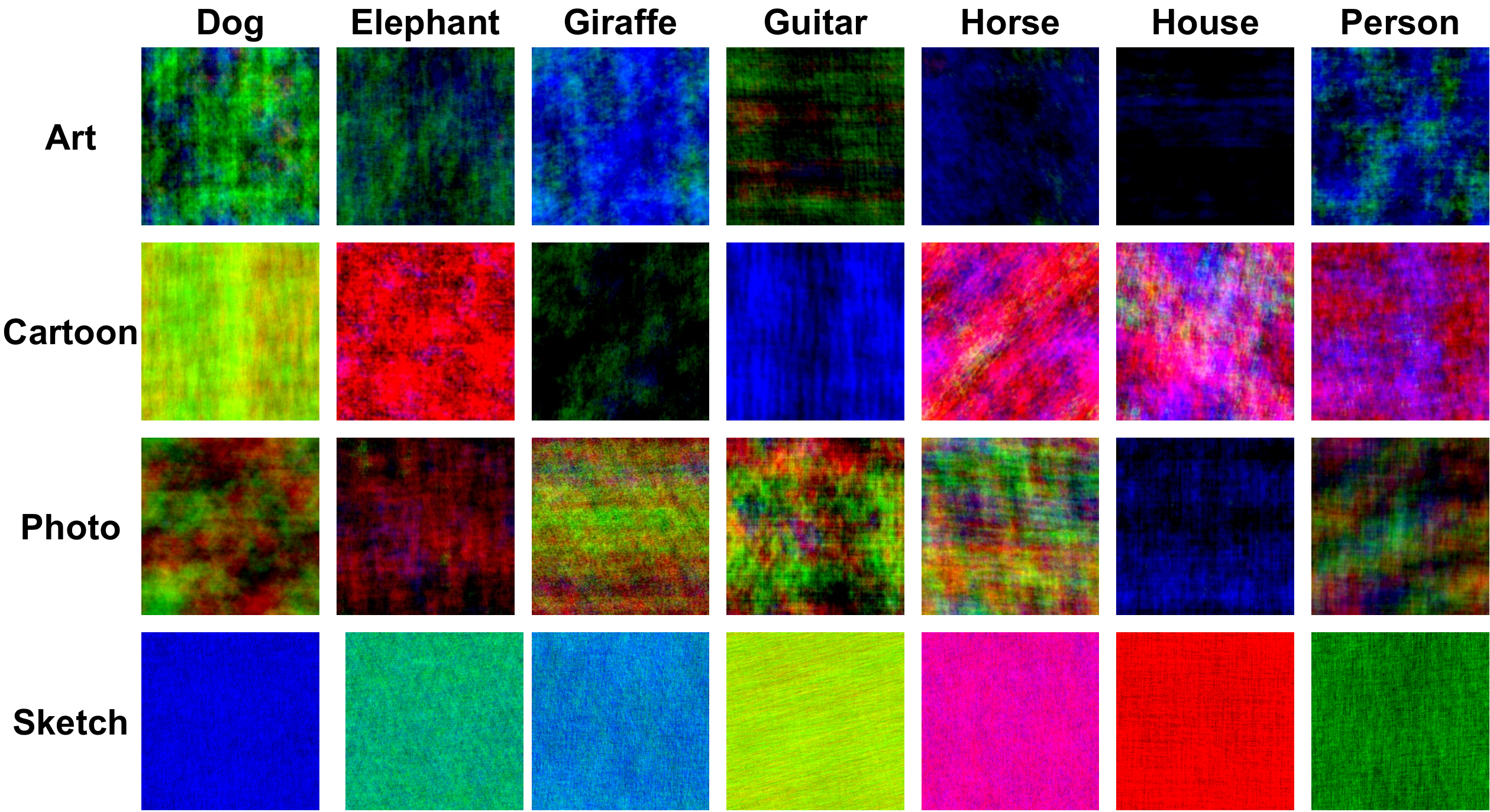}%
\label{fig:pacs_amp}}
\hfil
\subfloat[Phase Only.]{\includegraphics[width=0.49\linewidth]{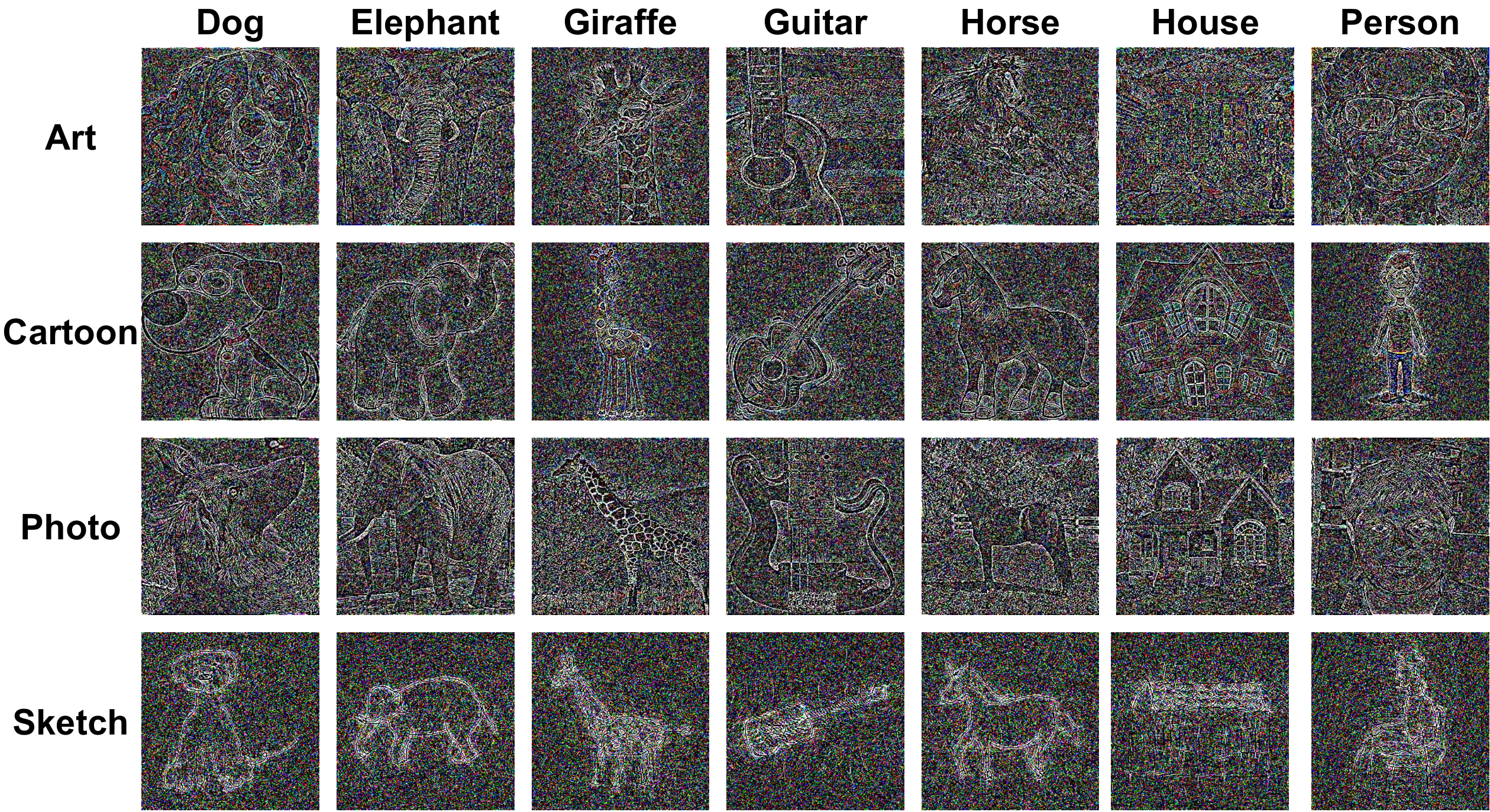}%
\label{fig:pacs_pha}}
\caption{Reconstructions for the Fourier components. We calculate the phase and amplitude spectrum for each image in Fig. \ref{fig:pacs}, and randomly initialize the phase (a) and amplitude (b) spectrum to a constant. (a) and (b) are reconstructed images with only amplitude information and phase information respectively.}
\label{fig:toy_exp}
\end{figure*}

For the phase spectrum, empirical experiments (Fig. \ref{fig:toy_exp}) show that the phase spectrum preserves highly structured spatial information in the phase-only reconstructed image, while the amplitude-only is entirely corrupted. Specifically, for each image patch from the pairs, the structural information (\textit{e.g.}, contour, edge) remains remarkably consistent, which makes significant contributions to recognition and positioning for the human visual system. Since the visual structures are the keys to describing different objects, learning from such information can facilitate the model to extract invariant semantics. Therefore, we assert that the secret to robust visual systems lies in the utilization of the phase spectrum and low sensitivity to the amplitude spectrum. Moreover, this consistent structural information presented in the image patches motivates us to establish spatial relationships for the phase spectrum. However, existing DG studies have largely overlooked this, which is a critical aspect of a generalizable representation.

In this paper, we seek to address DG problems from a frequency perspective. We analyze the distribution of the frequency components, and argue that \textit{domain shift is more likely to be caused by changes in the amplitude spectrum than the phase spectrum}. Inspired by this,  we propose \textit{Phase Matching} (PhaMa) to enhance generalizability against domain shifts, which perturbs the amplitude spectrum for adversarial training. Specifically, we randomly select two images from the source and mix their amplitude spectra through linear interpolation. Both the original images and the augmented versions are then fed into the network. Subsequently, we introduce a patch contrastive loss\cite{park2020contrastive} to encourage the matching of patch representations from the image pairs. This operation further alleviates the impact of the amplitude spectrum and establishes the spatial relationship of the phase spectrum, allowing the network to prioritize the phase spectrum of the image.

Our contributions can be summarized as follows: (1) We provide distribution analysis for the frequency components which clarifies the relationship with DG. (2) We introduce an effective algorithm called \textit{Phase Matching} (PhaMa), which guides the network to prioritize the phase spectrum for generalizable representation learning. (3) We provide a new state-of-the-art method on multiple domain generalization benchmarks. (4) Beyond empirical studies, we propose an intuitive causal view for domain generalization by specifying the causal/non-causal factors associated with the Fourier spectrum.

\section{Related Work}
\label{sec:relwo}

\noindent \textbf{Domain Generalization.} Domain generalization (DG) aims to train a model on source domains that perform well on unseen target domains. Data augmentation is a widely used technique to improve the generalization ability of models on out-of-distribution data\cite{wang2022generalizing,zhou2022domain}. MixUP\cite{zhang2018mixup} utilizes linear interpolations between two input samples and smooths the label. CutMix\cite{yun2019cutmix} generates training images by cutting and pasting from raw images. From the frequency perspective, recent works\cite{chen2021amplitude,xu2021fourier,lv2022causality} have incorporated the properties of phase and amplitude of the Fourier spectrum into DG\cite{xu2021fourier,lv2022causality} and Robustness\cite{chen2021amplitude}. Motivated by the observation that image style can be captured from latent feature statistics\cite{ulyanov2016instance,huang2017arbitrary}, many DG methods utilize adaptive instance normalization (AdaIN)\cite{huang2017arbitrary} and its variants\cite{nuriel2021permuted,zhou2021domain,li2022uncertainty} to synthesize novel feature statistics. Another way to tackle DG problems is domain-invariant representation learning. For example, MMD-AAE\cite{li2018domain} regularizes a multi-domain auto-encoder by minimizing the Maximum Mean Discrepancy (MMD) distance. \cite{zhao2020domain} minimizes the KL divergence between the conditional distributions of different training domains. DAL\cite{peng2019domain} disentangles domain-specific features using adversarial losses. 

\noindent \textbf{CNN Behaviors from Frequency Perspective.} Several frequency-based researches on CNN have been conducted. \cite{guo2018low} conducts adversarial attacks on low-frequency components and reveals that CNN primarily utilizes the low-frequency components for prediction. \cite{sharma2019effectiveness} demonstrates that CNN is vulnerable under low-frequency perturbations. On the other hand, \cite{wang2020high} observes that high-frequency components are essential for the generalization of CNN. Further, APR\cite{chen2021amplitude} provides a qualitative study for both the amplitude and phase spectrum, and argues that the phase spectrum is crucial for robust recognition. However, the spatial relationship of the phase spectrum remains unexplored, which we aim to investigate from a contrastive view.

\noindent \textbf{Contrastive Learning.} As a simple and powerful tool for visual representation learning, there has been a surge of impressive studies\cite{caron2021emerging,chen2020simple,he2020momentum,wu2018unsupervised} on contrastive learning\cite{hadsell2006dimensionality}. InstDisc\cite{wu2018unsupervised} uses a memory-bank to store the features for contrast. MoCo\cite{he2020momentum} builds a dictionary with a queue and adopts a momentum-update strategy for consistent representation learning. SimCLR\cite{chen2020simple} introduces a learnable projection head for better extraction. DINO\cite{caron2021emerging} trains the Vision Transformer (ViT)\cite{dosovitskiy2020vit} using contrastive learning, and gets explicit semantic information.

Given that contrastive learning only requires a definition of positive-negative pairs, it has recently been introduced into DG. PDEN\cite{li2021progressive} and SelfReg\cite{kim2021selfreg} aligns the cross-domain positive pairs by contrastive learning. PCL\cite{yao2022pcl} proposes a proxy-based contrast method for DG.
 
\section{Problem Definition}
\label{sec:pd}
Given a training set consisting of $M$ source domains $\mathcal{D}_{s}=\left\{ \mathcal{D}_{k}|k=1,\dots,M \right\}$ where $\mathcal{D}_{k}=\{(x_{l}^{k}, y_{l}^{k})\}_{l=1}^{n_{k}}$ denotes the $k$-th domain. The goal of domain generalization is to learn a robust and generalizable model $g:\mathcal{X}\to \mathcal{Y}$ from the $M$ source domains and achieve a minimum prediction error on the target domain $\mathcal{D}_{t}$, which is inaccessible during training:
\begin{equation}
	\min_{g} \mathbb{E}_{(x, y)\in \mathcal{D}_{t}}\left[\ell(g(x),y) \right].
\end{equation}
In this paper, we mainly consider an object recognition model $g(\cdot;\theta):\mathcal{X}\to \mathbb{R}^{N}$, where $\theta$ denotes the model parameters, $N$ is the number of categories in the target domain. Beyond simple classification tasks, we also evaluate our method on semantic segmentation task.

\section{Distribution Analysis for Frequency Components with Domain Generalization}
\label{sec:dis_aly}

We first clarify the relationship between frequency components and domain shifts as:
\begin{corollary}
   The amplitude spectrum tends to have huge variations under domain shifts, while the phase spectrum is relatively more domain-invariant.
   \label{cor:fc&ds}
\end{corollary}

Given a single-channel image $x$, its Fourier transform $\mathcal{F}(x)$ can be formulated as:
\begin{equation}
    \mathcal{F}(x)(u, v) =\iint_{\mathbb{R}^{2}} x(h, w) \cdot e^{-j2\pi(u h+v w)}.
    \label{eq:fft}
\end{equation}
The amplitude and phase spectrum are defined as:
\begin{equation}
\begin{aligned}
    \mathcal{A}&=\left[R^{2}(x)(u, v)+I^{2}(x)(u, v) \right]^{\frac{1}{2}}, \\
    \mathcal{P}&=\arctan\left[\frac{I(x)(u, v)}{R(x)(u, v)}\right],
\end{aligned} 
\label{eq:def_a&p}
\end{equation}
where $R(x)$ and $I(x)$ denote the real and imaginary part of $\mathcal{F}(x)$.The Fourier process can the be formulated as:
\begin{equation}
\begin{aligned}
    \mathcal{F}(x)(\mathcal{A}, \mathcal{P}) &= \mathcal{A} \times e^{-j \cdot \mathcal{P}} \\
    &= \mathcal{A} \times (\cos\mathcal{P}-j\cdot\sin\mathcal{P})
\end{aligned}
\label{eq:fft_a&p}
\end{equation}

Based on the observations in Sec. \ref{sec:intro}, we first make the following assumption \textbf{for the simplicity of derivation}:
\begin{assumption}
    The real and imaginary part of the complex spectrum follow the 2-D normal distribution.
\end{assumption}

Specifically, we consider the 2-D normal distribution $\mathcal{N}(0, \sigma^{2})$ for the complex spectrum:
\begin{equation}
    f(R, I) = \frac{1}{2\pi \sigma^{2}} e^{-\frac{R^{2}+I^{2}}{2\sigma^{2}}},
    \label{eq:2d-norm}
\end{equation}

With Eq. \eqref{eq:fft_a&p}, the distribution can be formulated as:
\begin{equation}
    \begin{aligned}
    f(\mathcal{F}(\mathcal{A}, \mathcal{P})) =& \mathcal{A} \times f(R=\mathcal{A}\cos\mathcal{P}, I =\mathcal{A}\sin\mathcal{P}) \\
    =& \mathcal{A} \times \frac{1}{2\pi \sigma^{2}} \cdot e^{-\frac{\mathcal{A}^{2}(\cos^{2}\mathcal{P}+\sin^{2}\mathcal{P})}{2\sigma^{2}}} \\
    =& \frac{\mathcal{A}}{\sigma^{2}}\cdot e^{-\frac{\mathcal{A}^{2}}{2\sigma^{2}} } \cdot \mathbb{I} \times \frac{1}{2\pi} \\
    =& \mathrm{Rayleigh}(\mathcal{A}\mid \sigma^{2})\cdot \mathrm{U}(\mathcal{P}\mid 0, 2\pi),
    \end{aligned}
\end{equation}
where $\mathcal{A} \ge 0$ and $0\le \mathcal{P} \le 2\pi$.

This form shows that the amplitude $\mathcal{A}$ can be arbitrarily large. Since the pixels are in the RGB Color Space $[0, 255]^{3}$, when domain shifts occur, the significant differences in the pixels can causes huge variations on the amplitude spectra between the cross-domain samples. On the other hand, the phase spectrum is restricted to a fixed range, which means that the impacts of domain shifts on the phase spectrum is quite weaker than the amplitude.

\section{Method}
\label{sec:method}
From the above observations, the amplitude spectrum is sensitive to domain shifts, while the phase spectrum is vital for recognition and less affected by domain shifts. Our hypothesis is that a robust representation remains \textit{invariant} to the phase spectrum of the object despite significant perturbations in the amplitude spectrum. Based on this motivation, we present Phase Matching as described next.

\subsection{Amplitude Perturbation Data Augmentation}
\label{sec:aada}
As discussed in Secs. \ref{sec:intro} and \ref{sec:dis_aly}, the amplitude spectrum usually has huge variations under domain shifts. The corresponding variations in RGB space thus undermine the generalizability of the DNN. To make the network robust to this perturbation, an intuitive way is to add attacks for the training examples to get adversarial gradients. Therefore, we introduce perturbations by linearly interpolating between the amplitude spectra of two randomly-sampled images, while maintaining the phase spectra unchanged as in\cite{xu2021fourier,lv2022causality}.

Formally, given an image $x\in \mathbb{R}^{H\times W\times 3}$, we can obtain the complex spectrum $\mathcal{F}\in \mathbb{C}^{H\times W\times 3}$ computed across the spatial dimension within each channel using FFT\cite{nussbaumer1981fast}:
\begin{equation}
	\mathcal{F}(x)(u, v) = \sum_{h=0}^{H-1}\sum_{w=0}^{W-1} x(h, w) \cdot e^{-j2\pi(\frac{h}{H}u+\frac{w}{W}v)}, 
\end{equation}
where $H$ and $W$ represent the height and width of the image respectively. 

We then perturb the amplitude spectra of two images $x_{o}$ and $ x_{o}'$, which are randomly selected from source domains, in the same way as MixUP \cite{zhang2018mixup}:
\begin{equation}
	\hat{\mathcal{A}}_{o}^{o'}=(1-\lambda) \mathcal{A}\left(x_{o}\right)+\lambda \mathcal{A}\left(x_{o}'\right),
	\label{eq:amp_mix}
\end{equation}
where $\lambda \sim U(0, \eta)$ and $\eta$ is a hyperparameter that controls the scale of perturbation. The phase-invariant image $x_{a}$ is then reconstructed from the combination of the original phase component and mixed amplitude component:
\begin{equation}
	x_{a} = \mathcal{F}^{-1}(\hat{\mathcal{A}}_{o}^{o'} \times e ^{-j \mathcal{P}(x_{o})}).
	\label{eq:ifft}
\end{equation}

The image pairs and the corresponding original labels are both fed to the model for training. The prediction loss is formulated as the standard Cross Entropy Loss:
\begin{equation}
	\mathcal{L}^{o(a)}_{cls} = -y^{T} \log (prob(x_{o(a)})),
\end{equation}
where $prob$ denotes the probability of each category. 

By utilizing this simple operation, we enhance the network's robustness against unknown amplitude shifts. More importantly, since the phase spectrum remains intact during this operation, we effectively create positive pairs, which paves the way for our subsequent contrastive regularization. This regularization further enhances the network's ability to capture phase spectrum features.

\begin{figure}[!t]
	\centering
	\includegraphics[width=0.95\linewidth]{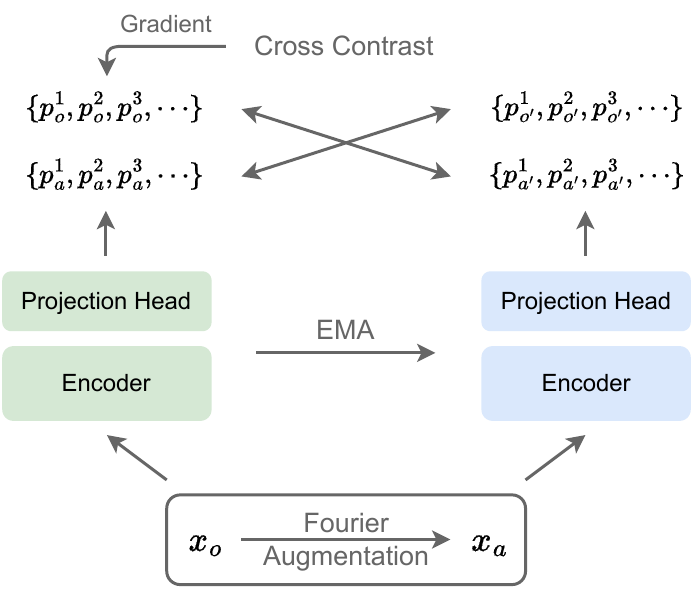}
	\caption{Framework of PhaMa. The Fourier-augmented image pairs ($x_{o}$ and $x_{a}$) are both fed to the encoder and the momentum-updated encoder. Then, the features from the last two layers are sent to a 2-layer nonlinear projection head for extracting patch representations $p_{o(a)}^{i}$. The cross-contrastive gradients are backpropagated for updating the encoder and projection head (in green); for the momentums (in blue), parameters are updated using EMA.}
	\label{fig:framework}
\end{figure}

\subsection{Matching Phase with Cross Patch Contrast} 
\label{sec:pm}

At the core of our method is \textit{matching} the phase spectrum of the original image with the corresponding augmented image; in other words, the representation of the image pairs should exhibit similarity or even be the same. Our research reveals that there is currently no neural network-based method that specifically focuses on extracting phase spectrum features. Consequently, we face challenges in directly matching the phase embeddings with common metrics such as L1, L2, and KLD. To address this, we incorporate contrastive learning \cite{hadsell2006dimensionality}, an unsupervised learning method that measures the similarities of sample pairs in a representation space, into our method.

Given the encoded hierarchal feature maps of an image $f_{t}(x)$, where $t$ denotes the index of the feature map, \textit{e.g.}, for a ResNet-like network, $t\in\left\{1, 2, 3, 4 \right\}$. Our choice of the feature representations is the \textit{last-two-levels} in a hierarchal network, in that the high-level features of the network are more likely to extract semantic-related information \cite{zeiler2014visualizing}. Specifically, for the two hierarchal representations $f_{3}\in \mathbb{R}^{C_{3}\times H_{3}\times W_{3}}$ and $f_{4}\in \mathbb{R}^{C_{4}\times H_{4}\times W_{4}}$, we resize $f_{4}$ using bilinear interpolation and concatenate them in the channel dimension, denoted as $z$, and send them to a 2-layer nonlinear projection head $p(z) = W^{(1)}\sigma(W^{(2)}z)$ as in\cite{chen2020simple}, where $\sigma$ denotes the ReLU activation function.

Inspired by the high consistency of the phase spectrum in preserving spatial structures, \textit{i.e.}, for the same position in the image pairs in Sec. \ref{sec:intro}, contours and edges are highly consistent. We aim to establish associations for each patch in the spatial dimension, \textit{i.e.}, make the patch representations from the same location \textit{similar}, and \textit{push away} those from different positions as far as possible. In this way, the encoded representations from each patch are consistent under the amplitude perturbations, and the network can learn from the invariant phase spectrum. Therefore, the following PatchNCE loss\cite{park2020contrastive} is considered:
\begin{equation}
		\mathcal{L}_{patch}^{o2a'}= -\sum_{i} \log \frac{\exp \left(p^{o}_{i}\cdot p^{a'}_{i} / \tau\right)}{\exp \left(p^{o}_{i}\cdot p^{a'}_{i} / \tau\right) + \sum_{j} \exp \left(p^{o}_{i}\cdot p^{a'}_{j} / \tau\right)},
		\label{eq:loss_match}	
\end{equation}
where $p^{o}$ and $p^{a'}$ are from the network (in green) and the momentum network (in blue), respectively. $i(j)\in\left\{1, \ldots, P \right\}$ denotes the index of the patch, and $\left(\cdot \right)$ denotes the inner product. $\tau$ is a temperature parameter. For the $i$-th patch in the original image $p^{o}_{i}$, patches in other locations in the augmented image $p^{a'}_{j}(j\ne i)$ are treated as negative samples. The contrast can then be set as a $P$-way classification problem. Algorithm \ref{alg:patchnce} provides the brief pseudo-code of this operation.

However, existing pretrained networks\cite{he2016deep,dosovitskiy2020vit} extract significantly different representations under huge perturbations of the amplitude spectrum, which tends to cause gradient collapse during the back-propagation in our experiments (\ref{sec:abl}). To alleviate the impact caused by amplitude perturbations, we propose the following two techniques:
\begin{itemize}
\item For both the original and augmented images, we adopt a momentum-updated rule to ensure consistent representation extraction. Following the approach proposed in\cite{he2020momentum}, we update the parameters of the network ($\theta_{n}$), and the momentum network's $\theta_{m}$, using the following rule:
\begin{equation}
    \theta_{m} \leftarrow m \theta_{m} + (1-m) \theta_{n}.
    \label{eq:moup}
\end{equation}
\item We perform the patch contrast operation (Eq. \eqref{eq:loss_match}) \textbf{\textit{across}} the patch representation of the original image and the augmented image from the network and the momentum network, respectively. The cross-contrastive loss is defined as:
\begin{equation}
    \mathcal{L}_{contr} = \mathcal{L}_{patch}^{o2a'} + \mathcal{L}_{patch}^{a2o'}.
\end{equation} 
\end{itemize}

\begin{algorithm}[t]
\caption{PyTorch-like pseudocode of patch contrast.} 
\begin{lstlisting}
# q, k: ori&aug features NxCxP
# t: temperature

# positive logits: NxPx1
l_pos = (q*k).sum(1)[:,:,None]
l_pos.norm()

# negative logits: NxPxP
l_neg = bmm(q.transpose(1,2), k)
l_neg.norm()

# remove diagonal entries
idt_mat = eye(P)[None,:,:]
l_neg.masked_fill_(idt_mat, -float('inf'))

# logits: NxPx(P+1) -> (N*P)x(P+1)
logits = cat([l_pos, l_neg], dim=2).flat(0,1)/t

# patch contrastive loss Eq.(9)
labels = zeros(N*P)
loss = CrossEntropyLoss(logits, labels)
\end{lstlisting}
\label{alg:patchnce}
\end{algorithm}


The overall objective function of our proposed method can be formulated as follows:
\begin{equation}
	\mathcal{L}_{PhaMa}=\frac{1}{2}(\mathcal{L}_{cls}^{o}+\mathcal{L}_{cls}^{a})+\beta \mathcal{L}_{contr},
	\label{eq:loss_total}
\end{equation}
where $\beta$ is a trade-off parameter.

\section{Experiments}
\label{sec:exp}

In this section, we perform experiments on various benchmarks to assess the effectiveness of our method in enhancing the network's generalization capabilities. The evaluation includes multi-domain classification, robustness against corruption, and semantic segmentation. We also provide ablation studies and visualization to better illustrate how our model operates and the extent of its impact.

\subsection{Multi-domain Classification}
\label{sec:mdc}

\subsubsection{Implementation Details}
\label{sec:mdc_imp}
\
\par
\textbf{Datasets.} We evaluate the generalization ability of our method on the following 3 datasets:
(1)\textbf{PACS}\cite{li2017deeper} is a commonly used benchmark for domain generalization, consisting of 9991 images from four distinct domains: Art Painting, Cartoon, Photo, Sketch; 
(2) \textbf{Digits-DG}\cite{zhou2020deep} consists of four datasets MNIST\cite{lecun1998gradient}, MNIST-M\cite{ganin2015unsupervised}, SVHN\cite{netzer2011reading}, SYN\cite{ganin2015unsupervised}; 
(3) \textbf{Office-Home}\cite{venkateswara2017deep} contains around 15,500 images of 65 categories from four domains: Artistic, Clipart, Product and Real World; 
(4) \textbf{VLCS}\cite{fang2013vlcs} includes images of different styles and origins from various data sources, such as Caltech, LabelMe, Pascal VOC 2007, and SUN09.

\textbf{Training.} For all DG benchmarks, we follow the leave-one-domain-out protocol with the official train-val split and report the classification accuracy (\%) on the entire held-out target domain. We also use standard augmentation, which consists of random resized cropping, horizontal flipping, and color jittering. For Digits-DG, all images are resized to $32\times 32$. We train the encoder (same as in \cite{zhou2020deep}) from scratch using SGD, batch size 64, and weight decay of 5e-4. The learning rate is initially 0.05 and is decayed by 0.1 every 20 epochs. For PACS and Office-Home, all images are resized to $224\times 224$. We use the ImageNet pretrained ResNet\cite{he2016deep} as the encoder and train the network with SGD, batch size 64, momentum 0.9, and weight decay 5e-4 for 50 epochs. The initial learning rate is 0.001 and is decayed by 0.1 at 80\% of the total epochs.

\textbf{Method-specific.} We use a sigmoid ramp-up\cite{tarvainen2017mean} for $\beta$ within the first 5 epochs in all experiments in this section. The scale parameter $\eta$ is set to 1.0 for Digits-DG, PACS, and VLCS 0.2 for Office-Home. The trade-off parameter $\beta$ is set to 0.1 for Digits-DG and 0.5 for PACS and Office-Home. The momentum $m$ is set to 0.9995. We also follow the common setting\cite{wu2018unsupervised} to let $\tau=0.07$.

\textbf{Model Selection.} We select the last-epoch checkpoint for evaluations on the target domain, and report the average results over 5 independent runs.

\subsubsection{Results Analysis}
\label{sec:mdc_ra}

\begin{table}[!t]
\centering
\caption{Leave-one-domain-out classification accuracy (\%) on PACS with ResNet pretrained on ImageNet. $\ddag$ denotes the reproduced results from FACT \cite{xu2021fourier}.}
\resizebox{\linewidth}{!}{
	\begin{tabular}{l|cccc|c}
		\toprule
		Method&  Art&  Cartoon&  Photo&  Sketch&  Avg.\\
		\midrule
		\multicolumn{6}{c}{ResNet-18} \\
		\midrule
		Baseline&  77.6&  76.7&  95.8&  69.5&  79.9 \\
		MixUP \cite{zhang2018mixup}&  76.8&  74.9&  95.8&  66.6&  78.5 \\
		CutMix \cite{yun2019cutmix}&  74.6&  71.8&  95.6&  65.3&  76.8 \\
		pAdaIN \cite{nuriel2021permuted}&  81.7&  76.6&  96.3&  75.1&  82.5 \\
		MixStyle \cite{zhou2021domain}&  82.3&  79.0&  96.3&  73.8&  82.8 \\
		DSU \cite{li2022uncertainty}&  83.6&  79.6&  95.8&  77.6&  84.1 \\
		MetaReg \cite{balaji2018metareg}&  83.7& 77.2& 95.5& 70.3& 81.7 \\
		JiGen \cite{carlucci2019domain}&  79.4&  75.2&  96.0&  71.3&  80.5 \\
		MASF \cite{dou2019domain}&  80.2&  77.1&  94.9&  71.6&  81.1 \\
		L2A-OT \cite{zhou2020learning}&  83.3&  78.0&  96.2&  73.6&  82.8 \\
		RSC$\ddag$  \cite{huang2020self}&  80.5&  78.6&  94.4&  76.0&  82.4 \\
		MatchDG \cite{mahajan2021domain}&  81.3&  \textbf{80.7}&  96.5&  79.7&  84.5 \\
		SelfReg \cite{kim2021selfreg}&  82.3&  78.4&  96.2&  77.4&  83.6 \\
		FACT \cite{xu2021fourier}& \textbf{85.3}&  78.3&  95.1&  79.1&  84.5 \\
		\midrule
		PhaMa (\textit{ours})&  84.9$\pm$0.5&  79.0$\pm$0.4&  \textbf{97.1}$\pm$0.3&  \textbf{79.7}$\pm$0.7&  \textbf{85.2} \\
		\midrule
		\multicolumn{6}{c}{ResNet-50} \\
		\midrule
		Baseline&  84.9&  76.9&  97.6&  76.7&  84.1 \\
		MetaReg \cite{balaji2018metareg}&  87.2&  79.2&  97.6&  70.3&  83.6 \\
		MASF \cite{dou2019domain}&  82.8&  80.4&  95.0&  72.2&  82.7 \\
		RSC$\ddag$  \cite{huang2020self}&  83.9&  79.5&  95.1&  82.2&  85.2 \\
		MatchDG \cite{mahajan2021domain}&  85.6&  82.1&  \textbf{97.9}&  78.7&  86.1 \\
		FACT \cite{xu2021fourier}&  \textbf{89.6}&  81.7&  96.7&  \textbf{84.4}&  88.1 \\
		\midrule
		PhaMa (\textit{ours})&  89.5$\pm$0.4&  \textbf{82.8}$\pm$0.5&  97.4$\pm$0.3&  84.0$\pm$0.6&  \textbf{88.4} \\
            \midrule
		\multicolumn{6}{c}{ViT-Small} \\
		\midrule
		Baseline&  89.1&  82.6&  98.9&  66.2&  84.2 \\
            MixStyle \cite{zhou2021domain}&  89.3&  82.9&  98.9&  65.1&  84.1 \\
		DSU \cite{li2022uncertainty}&  89.4&  83.1&  99.1&  66.3&  84.4 \\
		FACT \cite{xu2021fourier}&  90.6&  83.7&  99.1&  65.9&  84.8 \\
		\midrule
		PhaMa (\textit{ours})&  \textbf{91.7}$\pm$0.4&  \textbf{84.8}$\pm$0.4&  \textbf{99.4}$\pm$0.2&  \textbf{70.0}$\pm$0.5&  \textbf{86.5} \\
		\bottomrule
	\end{tabular}}
\label{tab:pacs}
\end{table}

\begin{table}[!t]
\centering
\caption{Leave-one-domain-out classification accuracy (\%) on Digits-DG with ConvNet in \cite{zhou2020deep}.}
\resizebox{\linewidth}{!}{
\begin{tabular}{l|cccc|c}
	\toprule
	Method&  MNIST&  MNIST-M&  SVHN&  SYN&  Avg. \\
	\midrule
	Baseline&  95.8&  58.8&  61.7&  78.6&  73.7 \\
	CCSA \cite{motiian2017unified}&  95.2&  58.2&  65.5&  79.1&  74.5 \\
	MMD-AAE \cite{li2018domain}&  96.5&  58.4&  65.0&  78.4&  74.6 \\
	CrossGrad \cite{shankar2018generalizing}&  96.7&  61.1&  65.3&  80.2&  75.8 \\
	DDAIG \cite{zhou2020deep}&  96.6&  \textbf{64.1}&  68.6&  81.0&  77.6 \\
	Jigen \cite{carlucci2019domain}&  96.5&  61.4&  63.7&  74.0&  73.9 \\
	L2A-OT \cite{zhou2020learning}&  96.7&  63.9&  68.6&  83.2&  78.1 \\
	MixStyle \cite{zhou2021domain}&  96.5&  63.5&  64.7&  81.2&  76.5 \\
	FACT \cite{xu2021fourier}&  96.8&  63.2  &\textbf{73.6}  &89.3  &80.7  \\
	\midrule
	PhaMa (\textit{ours})&  \textbf{97.4}$\pm$0.2&  64.0$\pm$0.2&  73.4$\pm$0.4&  \textbf{90.5}$\pm$0.4&  \textbf{81.3} \\
	\bottomrule
\end{tabular}}
\label{tab:digits_dg}
\end{table}

\begin{table}[!t]
\centering
\caption{Leave-one-domain-out classification accuracy (\%) on Office-Home with ResNet-18 pretrained on ImageNet.}
\resizebox{\linewidth}{!}{
\begin{tabular}{l|cccc|c}
	\toprule
	Method&  Art&  Clipart&  Product&  Real&  Avg.\\
	\midrule
	Baseline&  57.8&  52.7&  73.5&  74.8&  64.7 \\
	CCSA \cite{motiian2017unified}&  59.9&  49.9&  74.1&  75.7&  64.9 \\
	MMD-AAE \cite{li2018domain}&  56.5&  47.3&  72.1&  74.8&  62.7 \\
	CrossGrad \cite{shankar2018generalizing}&  58.4&  49.4&  73.9&  75.8&  64.4 \\
	DDAIG \cite{zhou2020deep}&  59.2&  52.3&  74.6&  76.0&  65.5 \\
	L2A-OT \cite{zhou2020learning}&  \textbf{60.6}&  50.1&  74.8&  \textbf{77.0}&  65.6 \\
	Jigen \cite{carlucci2019domain}&  53.0&  47.5&  71.4&  72.7&  61.2 \\
	MixStyle \cite{zhou2021domain}&  58.7&  53.4&  74.2&  75.9&  65.5 \\
	DSU \cite{li2022uncertainty}&  60.2&  \textbf{54.8}&  74.1&  75.1&  66.1 \\
	FACT \cite{xu2021fourier}&  60.3&  \textbf{54.8}&  74.4&  76.5&  66.5 \\
	\midrule
	PhaMa (\textit{ours})&  60.3$\pm$0.4&  54.2$\pm$0.3&  \textbf{75.4}$\pm$0.6&  76.4$\pm$0.6&  \textbf{66.6}  \\
	\bottomrule
\end{tabular}}
\label{tab:office}
\end{table}

\begin{table}[!t]
\centering
\caption{Leave-one-domain-out classification accuracy (\%) on VLCS with ResNet-18 pretrained on ImageNet.}
\begin{tabular}{l|cccc|c}
	\toprule
	Method&  Art&  Clipart&  Product&  Real&  Avg.\\
	\midrule
	Baseline&  91.8&  61.8&  67.5&  68.8&  72.5 \\
	JiGen \cite{carlucci2019domain}&  96.2&  62.1&  70.9&  71.4&   75.1 \\
	M-ADA&  74.3&  48.3&  45.3&  33.8&  50.5 \\    
	DG-MMLD &  59.2&  52.3&  74.6&  76.0&  65.5 \\
	StableNet \cite{zhang2021deep}&  \textbf{96.7}&  65.4&  73.4&  74.5&  77.6 \\
	\midrule
	PhaMa&   96.3$\pm$0.3&  \textbf{65.9$\pm$0.4}&  \textbf{75.2$\pm$0.2}&  \textbf{75.7$\pm$0.3}&  \textbf{78.3} \\
	\bottomrule
\end{tabular}
\label{tab:vlcs}
\end{table}
\
\par
\textbf{PACS.} The experimental results presented in Tab. \ref{tab:pacs} show a significant improvement of PhaMa compared to the baseline approach. Particularly, PhaMa exhibits substantial improvements in average accuracy, notably in the Art, Cartoon, and Sketch domains. It is worth noting that previous DG methods experienced a slight drop in accuracy for the Photo domain, which shares similar domain characteristics with the ImageNet dataset, and this drop might be attributed to ImageNet pretraining. However, our method is capable of maintaining, or even improving the performance in the Photo domain, demonstrating that contrasting phase components does not degrade raw representations.

Note that other contrast-based methods (e.g., MatchDG and SelfReg) also demonstrate competitive performance on Photo and other domains. This observation highlights the effectiveness of contrastive learning as a powerful tool for DG. Meanwhile, the exceptional performance of PhaMa suggests that introducing spatial contrast for the phase spectrum can yield promising results.

In addition to CNN networks, we have also conducted experiments utilizing the Vision Transformer (ViT) \cite{dosovitskiy2021an} as the encoder network. Our approach demonstrates competitive performance against Baseline and FACT\cite{xu2021fourier}, especially on Sketch domain. Since ViT also packs the image into patches, our method can be adapted to this stream of networks easily. The superior performance indicates the importance of establishing spatial relationships through patch representation for Domain Generalization (DG).  It is noteworthy that the ImageNet-21k pretrained weights were employed in this setting, making even marginal improvements over prior arts a challenging endeavor.

\textbf{Digits-DG.} Results are shown in Tab. \ref{tab:digits_dg}. PhaMa achieves significant improvement over the baseline method and surpasses previous domain-invariant methods by a large margin. Since our method adopts the same augmentation technique as FACT\cite{xu2021fourier}, we compare our method with it. Our method shows slight improvement over FACT, indicating that matching the patch feature is more suitable for cross-domain learning.

\textbf{Office-Home.} Results are shown in Tab. \ref{tab:office}. It can be observed that our method brings obvious improvement over the baseline method and also achieves competitive results against FACT. By introducing amplitude perturbations and the patch contrastive loss, the model can learn to alleviate the amplitude impacts and focus on the phase spectrum.

\textbf{VLCS.} The results are shown in Tab. \ref{tab:vlcs}. PhaMa achieves state-of-the-art performance on VLCS with the highest average accuracy. Our method surpasses other methods on 3 domains with a large margin.

\subsection{Robustness Against Corruptions}
\label{sec:rtc}

\subsubsection{Implementation Details}
\label{sec:rtc_imp}
\
\par
\textbf{Datasets.} We evaluate the robustness against corruptions of our method on CIFAR-10-C, CIFAR-100-C\cite{hendrycks2019robustness}. The two datasets are constructed by corrupting the test split of original CIFAR datasets with a total of 15 corruption types (\textit{noise, blur, weather,} and \textit{digital}). Note that the 15 corruptions are not introduced during training.

\textbf{Training.} Following\cite{chen2021amplitude}, we report the mean Corruption Error (\%) for various networks, including ResNet-18\cite{he2016deep}, 40-2 Wide-ResNet\cite{zagoruyko2016wide}, DenseNet-BC ($k=2, d=100$)\cite{huang2017densely}, and ResNeXt-29 ($32\times4$)\cite{xie2017aggregated}. All networks use an initial learning rate of 0.1 which decays every 60 epochs. We train all models from scratch for 200 epochs using SGD, batch size 128, momentum 0.9. All input images are randomly processed with resized cropping and horizontal flipping.

\begin{table}[!t]
    \caption{Mean Classification Error (\%) on CIFAR-10(100)-C.}
    \resizebox{\linewidth}{!}{
    \begin{tabular}{l|cccc}
		\toprule
		Method&  ResNet&  DenseNet&  WideResNet&  ResNeXt\\
		\midrule
		\multicolumn{5}{c}{CIFAR-10-C} \\
		\midrule
		Standard&  -&  30.7&  26.9&  27.5 \\
		Cutout \cite{devries2017improved}&  -&  32.1&  26.8&  28.9 \\
		MixUP \cite{zhang2018mixup}&  -&  24.6&  22.3&  22.6 \\
		CutMix \cite{yun2019cutmix}&  -&  33.5&  27.1&  29.5 \\
		Adv Training&  -&  27.6&  26.2&  27.0 \\
		APR \cite{chen2021amplitude}&  \textbf{16.7}&  20.3&  18.3&  18.5 \\
		\midrule
		PhaMa (\textit{ours})&  17.3$\pm$0.4&  \textbf{20.2}$\pm$0.4&  \textbf{17.5}$\pm$0.3&  \textbf{18.0}$\pm$0.2 \\
		\midrule
		\multicolumn{5}{c}{CIFAR-100-C} \\
		\midrule
		Standard&  -&  59.3&  53.3&  53.4 \\
		Cutout \cite{devries2017improved}&  -&  59.6&  53.5&  54.6 \\
		MixUP \cite{zhang2018mixup}&  -&  55.4&  50.4&  51.4 \\
		CutMix \cite{yun2019cutmix}&  -&  59.2&  52.9&  54.1 \\
		Adv Training&  -&  55.2&  55.1&  54.4 \\
		APR \cite{chen2021amplitude}&  43.8&  49.8&  44.7&  44.2 \\
		\midrule
		PhaMa (\textit{ours})&  \textbf{43.5}$\pm$0.5&  \textbf{48.3}$\pm$0.4&  \textbf{43.6}$\pm$0.3&  \textbf{41.1}$\pm$0.4 \\
		\bottomrule
    \end{tabular}}
    \label{tab:cifar}
\end{table}

\textbf{Method-specific.} All configurations are consistent with Sec. \ref{sec:mdc_imp} except for the sigmoid ramp-up.

\textbf{Model Selection.} We select the last-epoch checkpoint for evaluations on the corrupted dataset, and report the average results over 5 independent runs.

\subsubsection{Results Analysis}
\label{sec:rtc_ra}

Results on CIFAR-10-C and CIFAR-100-C are shown in \ref{tab:cifar}. PhaMa outperforms conventional data augmentation techniques (Cutout, MixUP, CutMix) by a significant margin. Furthermore, PhaMa shows slight improvements over APR, which also employs a technique similar to ours in the frequency domain. This observation further highlights the significance of establishing spatial relationships within the phase spectrum, leading to a more effective extraction of intrinsic representations.

\subsection{Semantic Segmentation}
\label{sec:semseg}

\begin{table}[!t]
\centering
\caption{Results on GTA5$\rightarrow$Cityscapes.}
    \begin{tabular}{c|cc}
    \toprule
    Method&  mIoU(\%)&  mAcc(\%)  \\
    \midrule
    Baseline&  37.0&   51.5\\ 
    p-AdaIN \cite{nuriel2021permuted}&  38.3 & 52.1  \\
    MixStyle \cite{zhou2021domain}&  40.3  &53.8 \\
    DSU \cite{li2022uncertainty}& 43.1 &57.0\\
    PhaMa& \textbf{44.7$\pm$0.4}&  \textbf{58.1$\pm$0.3}\\
    \bottomrule
    \end{tabular}
\label{tab:seg}
\end{table}

\subsubsection{Implementation Details}
\
\par
\textbf{Datasets.} GTA5 \cite{richter2016playing} is a synthetic dataset generated from Grand Theft Auto 5 game engine, while Cityscapes \cite{cordts2016cityscapes} is a real-world dataset collected from different cities in Germany.

\textbf{Training.} Configurations are consistent with \cite{li2022uncertainty} using DeepLab-v2\cite{chen2017deeplab} with ResNet-101 as the backbone network. The model is trained with 20000 iterations. We report the mean Intersection over Union (mIoU) and mean Accuracy (mAcc) of all object categories.

\textbf{Model Selection.} We select the last-iteration checkpoint for
evaluations on the GTA5 dataset, and report the average
results over 2 independent runs.

\subsubsection{Results Analysis}

Results are shown in Tab. \ref{tab:seg}. For pixel-level classification tasks, the performance can be hindered by distribution shifts. Our method is robust to the shifts in different driver scenes. By establishing spatial relationships between image patches, our method can better localize objects and surpass others on mIoU.

\subsection{Ablation Studies}
\label{sec:abl}

\subsubsection{Effects of Different Modules} We conduct ablation studies to investigate the impact of each module in our method in Sec. \ref{abl:module}. Compared with the baseline, the amplitude perturbation data augmentation module (APDA) plays a significant role in our method, lifting the performance by a margin of 4.5\%. We exclude the momentum-updated encoder (MoEnc) for variant B, and the performance drops by nearly 1\%, demonstrating the over-dependence on the amplitude spectrum makes the feature extraction inefficient. With the cross-contrast operation for the image pairs, PhaMa surpasses variants C and D, suggesting that keeping the consistency between the image pairs is important for the training of contrastive learning.

\subsubsection{Sensitivity of Trade-off Parameter} 
\begin{figure}[!t]
    \centering
    \includegraphics[width=0.95\linewidth]{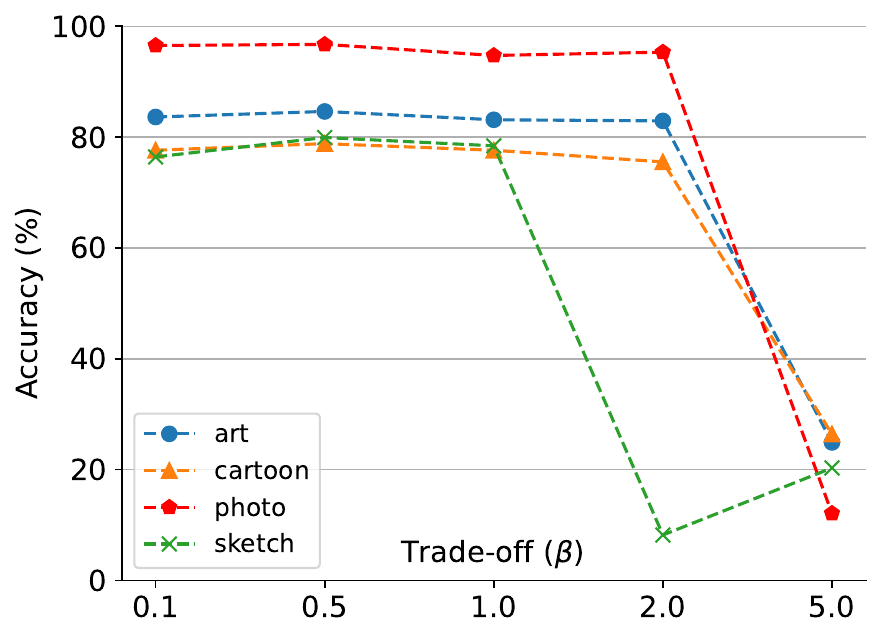}
    \caption{Evaluation of different trade-off parameters.}
    \label{fig:abl_beta}
\end{figure}

The hyperparameter $\beta$ controls the trade-off between the classification loss and the patch contrastive loss. We experiment with different values of $\beta$ from the set $\left\{0.1, 0.5, 1.0, 2.0, 5.0 \right\}$. The results, depicted in Fig. \ref{fig:abl_beta}, show that for small $\beta$ values, there are slight oscillations. However, when $\beta$ is set to a large value, the training process collapses. The collapse might be attributed to an over-dependence on the amplitude spectrum of the ImageNet pretrained weights. As $\beta$ increases, the intense contrast may likely corrupt the raw representation.

\begin{table*}[!t]
    \centering
    \caption{Effects of different modules on PACS with ResNet-18.}
    \resizebox{0.8\linewidth}{!}{
    \begin{tabular}{l|cccc|cccc|c}
		\toprule
		Method&  APDA&  $\mathcal{L}_{patch}^{o2a}$&  $\mathcal{L}_{patch}^{a2o}$&  MoEnc&  Art&  Cartoon&  Photo&  Sketch&  Avg. \\
		\midrule
		Baseline& -&  -&  -&  -&  77.6&  76.7&  95.8&  69.5&  79.9 \\
		\midrule
		Variant A& \Checkmark&  -&  -&  -&  83.9&  76.9&  95.5&  77.6&  83.4 \\
		Variant B& \Checkmark&  \Checkmark&  \Checkmark&  -&  83.2&  77.1&  95.5&  79.0&  83.7 \\
		Variant C& \Checkmark&  \Checkmark&  -&  \Checkmark&  84.2&  78.7&  96.1&  79.5&  84.6 \\
		Variant D& \Checkmark&  -&  \Checkmark&  \Checkmark&  84.1&  78.4&  95.5&  79.1&  84.5 \\
		\midrule
		PhaMa&  \Checkmark&  \Checkmark&  \Checkmark&  \Checkmark&  \textbf{84.8}&  \textbf{79.1}&  \textbf{96.6}&  \textbf{79.7}&  \textbf{85.1} \\
		\bottomrule
    \end{tabular}}
    \label{abl:module}
\end{table*}

\subsubsection{Choices of Matching Loss} Since our objective is to match the representation of each patch from the image pairs, we evaluate various types of matching loss, including SmoothL1, MSE, and PatchNCE. As demonstrated in Tab. \ref{tab:abl_loss}, the PatchNCE loss outperforms the others on Art Painting, Cartoon, and Sketch domains, while also showing competitive performance on the Photo domain. The inferior performance of SmoothL1 and MSE is probably due to the simplistic alignment of the representation, which lacks focus on discriminating the positive patch from the negatives. This phenomenon corroborates the observations in Sec. \ref{sec:intro}, that the spatial relationship carried in the phase information is crucial to image recognition.

\begin{table}[!t]
    \centering
    \caption{Evaluation of different types of matching loss.}
    \resizebox{0.9\linewidth}{!}{
	\begin{tabular}{c|cccc|c}
			\toprule
			Type&  Art&  Cartoon&  Photo&  Sketch&  Avg.\\
			\midrule
			SmoothL1&  83.7&  76.6&  96.4&  72.3&  82.3 \\
			MSE&  81.8&  77.4&  96.6&  76.1&  83.0 \\
			PatchNCE&  84.8&  79.1&  96.6&  79.7&  85.1 \\
			\bottomrule
    \end{tabular}}
    \label{tab:abl_loss}
\end{table}

\subsubsection{Impacts of Hierarchies} Table \ref{tab:abl_position} compares different hierarchical positions (indexed with 1-4) in the ResNet blocks for contrastive learning. As can be seen, the representation extracted from the last two layers occupies the least GPU memory while achieving the best effective performance.

\begin{table}[!t]
    \centering
    \caption{Average Accuracy (\%) and GPU Memory-Usage (GB) of different posiotions.}
    \resizebox{0.6\linewidth}{!}{
    \begin{tabular}{c|c|c}
	\toprule
	Position&  Avg Acc&  Mem-Usage \\
	\midrule
	1, 2&  82.2&  $\approx$43.3 \\
	2, 3&  83.5&  $\approx$12.1 \\
	3, 4&  85.1&  $\approx$11.8 \\
	\bottomrule
    \end{tabular}}
    \label{tab:abl_position}
\end{table}

\subsection{Visualization}
\subsubsection{t-SNE Distribution}
\begin{figure}[!t]
\centering
\subfloat[Baseline.]{\includegraphics[width=0.49\linewidth]{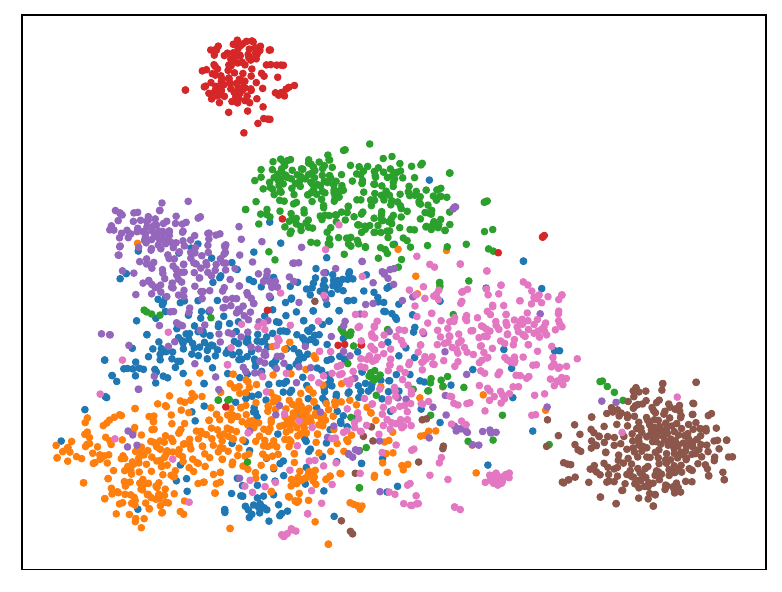}%
\label{fig:vis_base}}
\hfil
\subfloat[PhaMa.]{\includegraphics[width=0.49\linewidth]{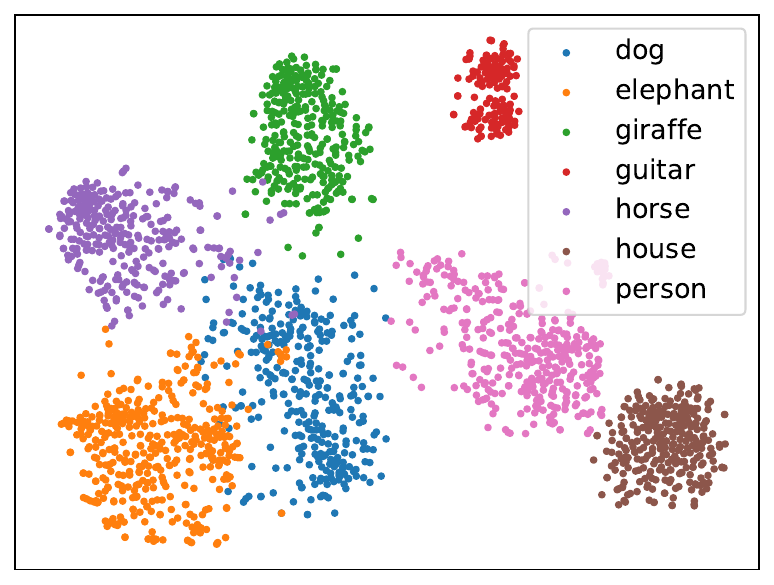}%
\label{fig:vis_phama}}
\caption{t-SNE visualization on Baseline and PhaMa. We visualize the flattened last-layer embeddings of ResNet-18 on PACS with Art, Photo and Sketch as the source domains, and Cartoon as the target domain.}
\label{fig:vis}
\end{figure}

To intuitively present PhaMa's effects on feature representations, we visualize the feature representation vectors of different categories in the unseen domain using t-SNE \cite{van2008visualizing} in Fig. \ref{fig:vis}. Compared with the baseline method, features from the same category become more compact with our method. The clustered representations illustrate that our method can alleviate perturbations caused by domain shifts and extract more domain-invariant features.

\subsubsection{Attention Maps}
\begin{figure*}[!t]
	\centering
	\includegraphics[width=\linewidth]{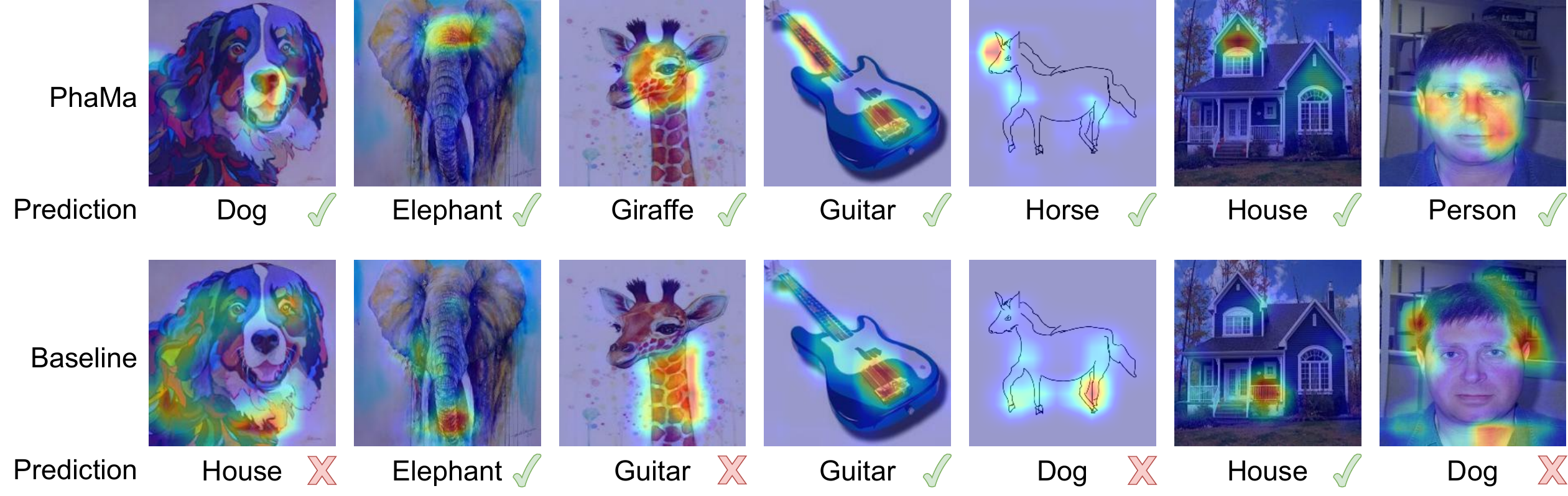}
	\caption{Visualization of attention maps on PACS. We each use the specific leave-on-domain weights to visualize the attention maps using Grad-CAM \cite{selvaraju2017grad}.}
	\label{fig:pacs_cam_pred}
\end{figure*}

To visually verify the claim that the representations learned by PhaMa can prioritize the phase spectrum of the image, we present the attention maps of the baseline method and PhaMa using Grad-CAM \cite{selvaraju2017grad}. As shown in Fig. \ref{fig:pacs_cam_pred}, the representations learned by PhaMa focus more on category-related information, thus verifying the effectiveness of our method in extracting domain-variant representations.

\section{Statistical Analysis of the Amplitude Spectrum for Cross-Domain Samples}
\label{app:sta_aly}

In this section, we first provide the definition of two commonly used frequency domain statistics and present the results on the amplitude spectra from PACS \cite{li2017deeper}, Digits-DG \cite{zhou2020deep}, Office-Home \cite{venkateswara2017deep} and CIFAR-10(100)(-C) \cite{hendrycks2019robustness}.

\subsection{Mathematical Definition}

To obtain the statistical features of the amplitude spectra, we calculate two commonly-used frequency statistics for the amplitude spectra of the selected samples.

\vskip 5pt
\noindent \textbf{Centroid Frequency} represents the center or balance point of the signal energy, which is the weighted average of the spectrum. In our experiments, it can be defined as:
\begin{equation}
	F_{c} = \frac{\sum{X_{i} \cdot X_{i}^{2}}}{\sum{X_{i}^{2}}},
\end{equation}
where $X_{i}$ denotes the amplitude of the $i$-th frequency component.

\vskip 5pt
\noindent \textbf{Frequency Standard Deviation} is a statistical measure used in the frequency domain to quantify the breadth or dispersion of signal frequency distribution, which can be formulated as:
\begin{equation}
	F_{std} = \sqrt{\frac{\sum{(X_{i}-F_{c})^{2} \cdot X_{i}^{2}}}{\sum{X_{i}^{2}}}}.
\end{equation}

\subsection{Comparison on DG Benchmarks}

\begin{figure}[!t]
    \centering
    \subfloat[$F_{c}$.]{
        \includegraphics[width=0.47\linewidth]{figs/pacs_amp_box_fc.pdf}
        \label{fig:pacs_box_fc}}
    \hfill
    \subfloat[$F_{std}$.]{
        \includegraphics[width=0.47\linewidth]{figs/pacs_amp_box_fstd.pdf}
        \label{fig:pacs_box_fstd}}

    \subfloat[t-SNE.]{
        \includegraphics[width=0.8\linewidth]{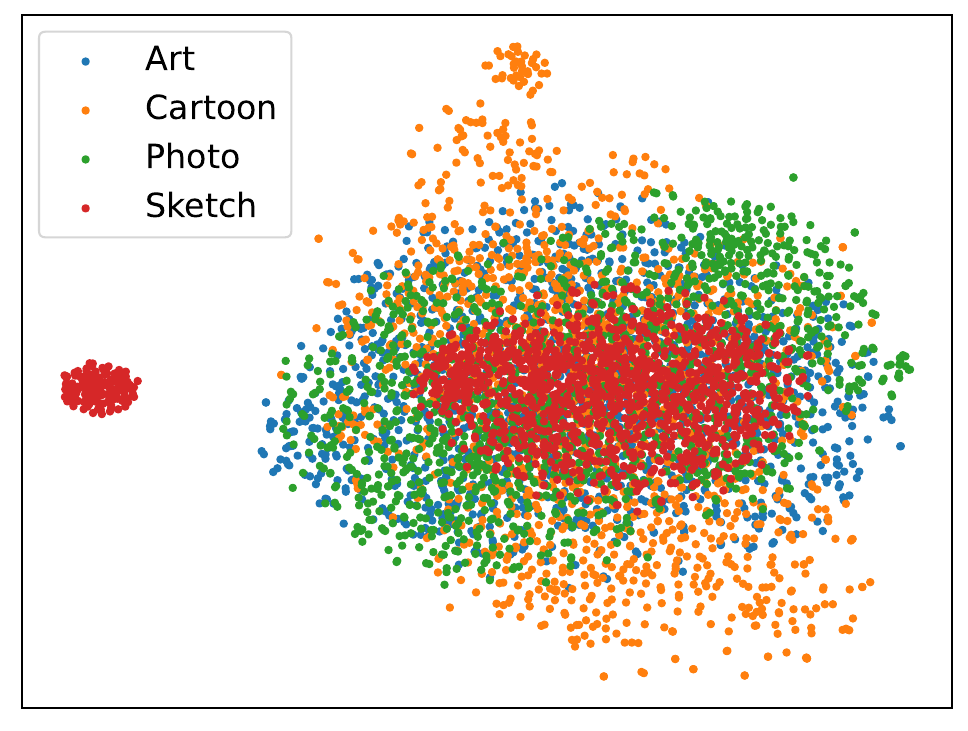}
        \label{fig:pacs_tsne}}
    \caption{(a) and (b) are the boxplots of centroid frequency and frequency standard deviation of the amplitude spectra from PACS. (c) is the t-SNE visualization of the amplitude spectra.}
    \label{fig:pacs_amp_sta}
\end{figure}

\noindent \textbf{PACS.} We randomly select 1500 images from each domain in PACS, and calculate the in-domain centroid frequency $F_{c}$ and frequency standard deviation $F_{std}$ for the amplitude spectra. As shown in Fig. \ref{fig:pacs_amp_sta}, it is clear that each domain's amplitude spectra exhibit a distinct distribution. For Art Painting and Photo, their distributions are similar, whereas Cartoon and Sketch show significant statistical differences. These results support the observations made in the Introduction and further reinforce the susceptibility of the amplitude spectrum to domain shifts.

\begin{figure}[!t]
    \centering
    \subfloat[$F_{c}$.]{
        \includegraphics[width=0.475\linewidth]{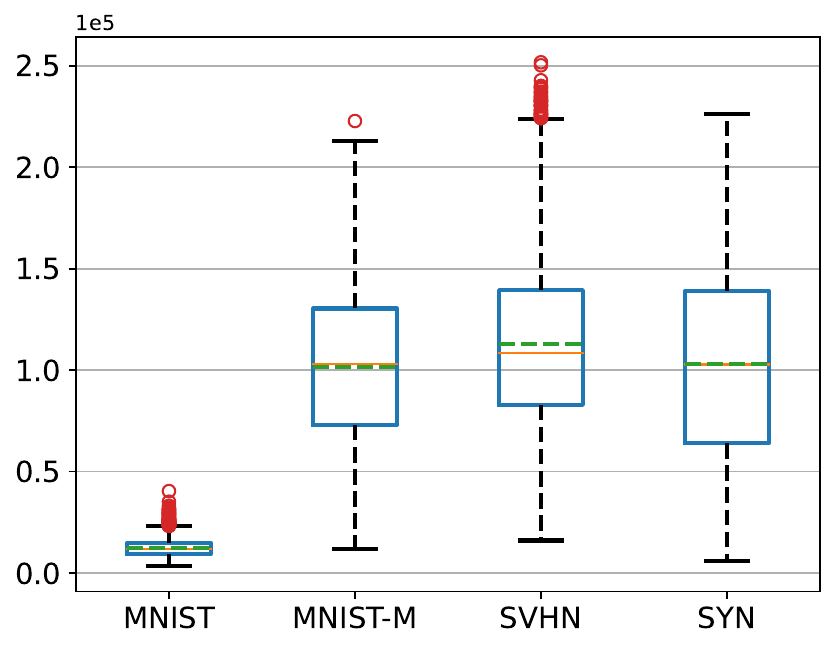}
        \label{fig:digits_box_fc}}
    \hfill
    \subfloat[$F_{std}$.]{
        \includegraphics[width=0.465\linewidth]{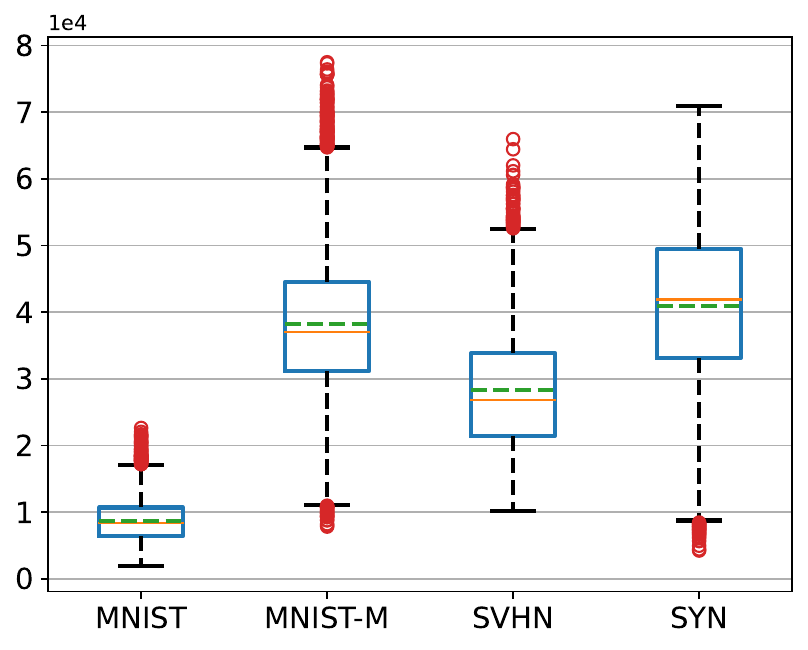}
        \label{fig:digits_box_fstd}}

    \subfloat[t-SNE.]{
        \includegraphics[width=0.8\linewidth]{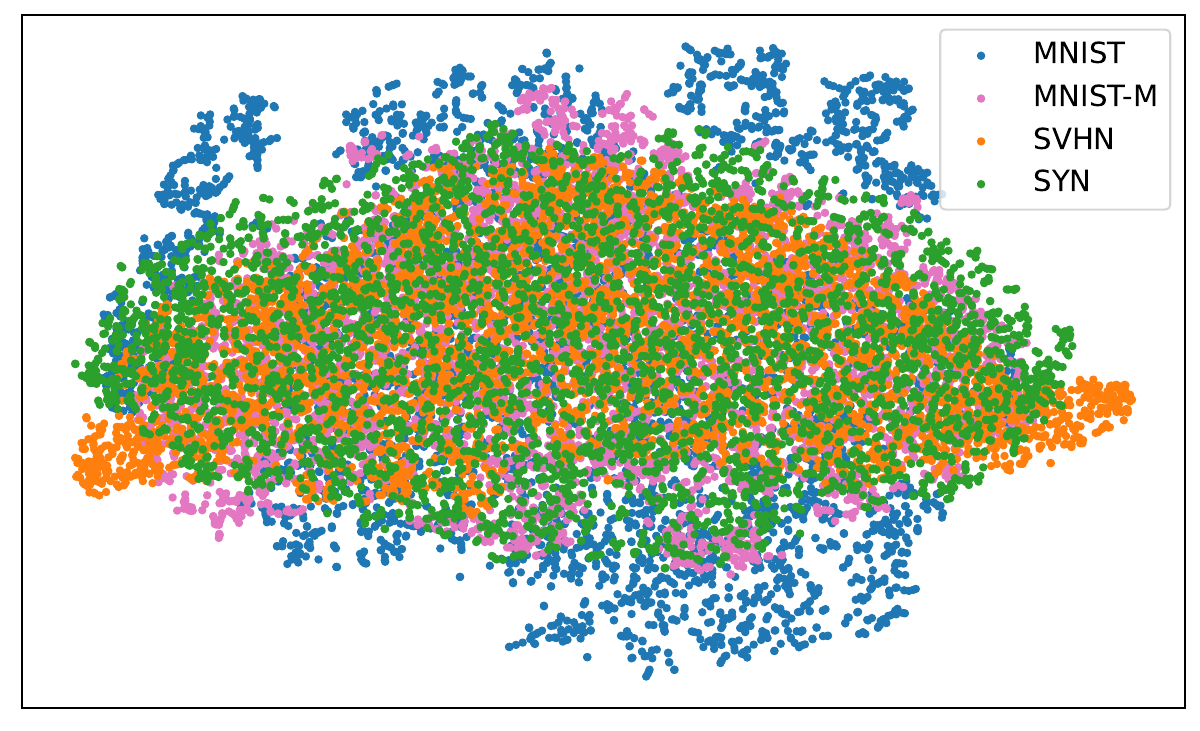}
        \label{fig:digits_tsne}}
    \caption{(a) and (b) are the boxplots of centroid frequency and frequency standard deviation of the amplitude spectra from Digits-DG. (c) is the t-SNE visualization of the amplitude spectra.}
    \label{fig:digits_amp}
\end{figure}

\begin{figure}[!t]
    \centering
    \subfloat[$F_{c}$.]{
        \includegraphics[width=0.475\linewidth]{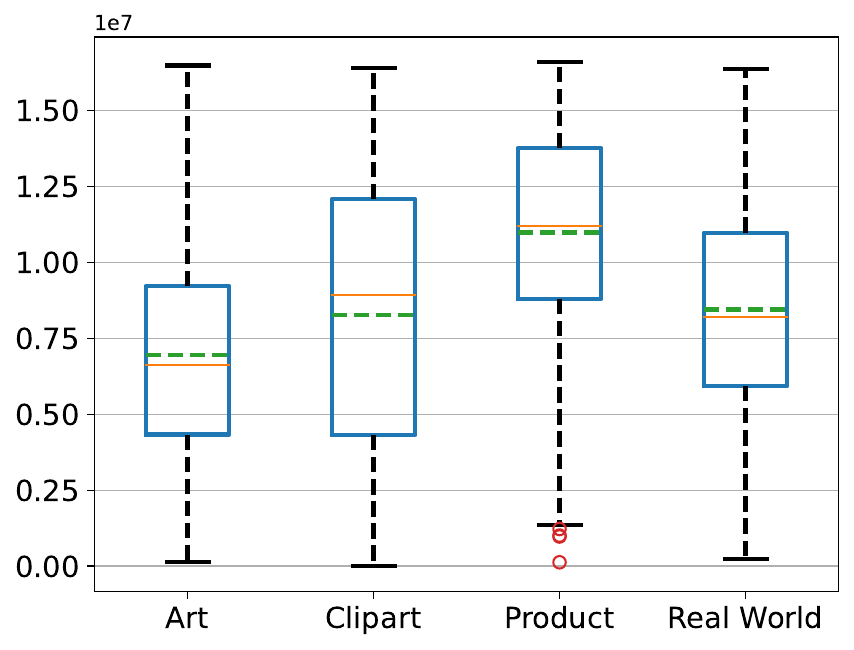}
        \label{fig:oh_box_fc}}
    \hfill
    \subfloat[$F_{std}$.]{
        \includegraphics[width=0.45\linewidth]{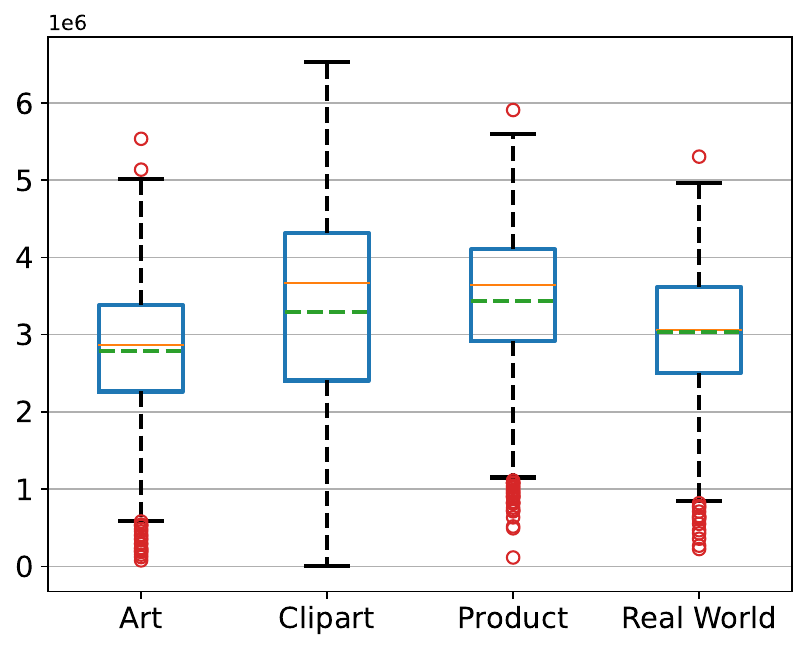}
        \label{fig:oh_box_fstd}}

    \subfloat[t-SNE.]{
        \includegraphics[width=0.8\linewidth]{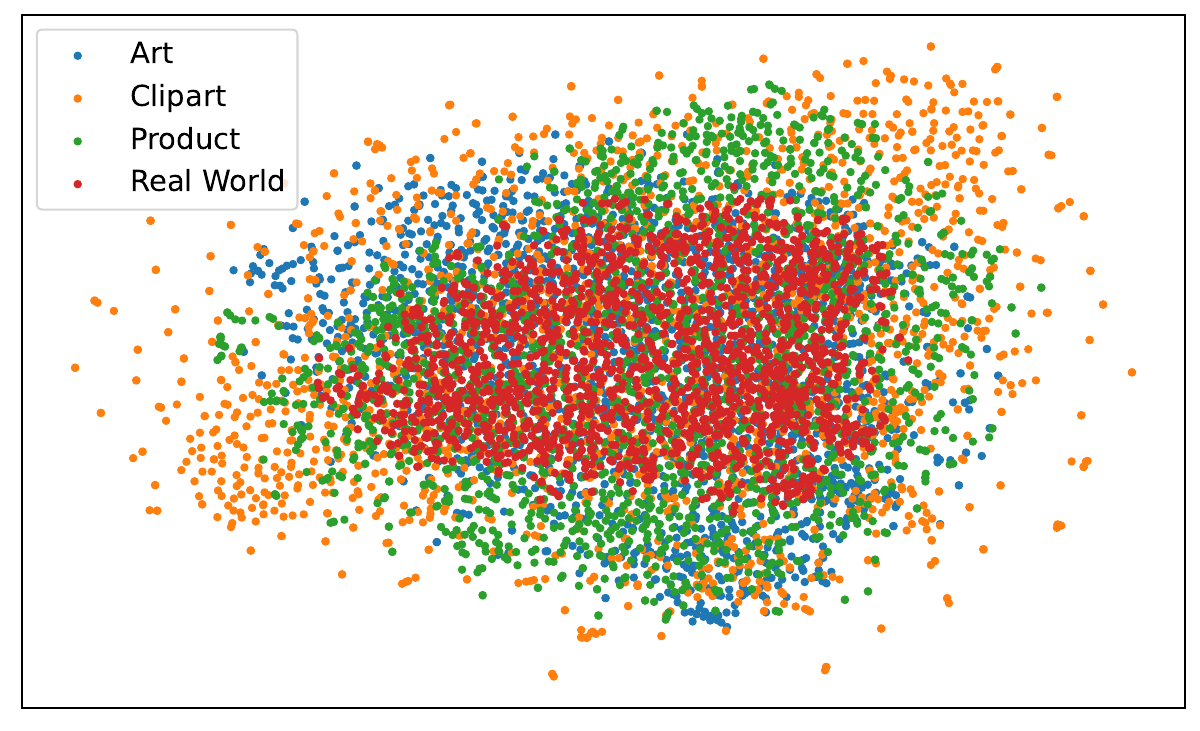}
        \label{fig:oh_tsne}}
    \caption{(a) and (b) are the boxplots of centroid frequency and frequency standard deviation of the amplitude spectra from Office-Home. (c) is the t-SNE visualization of the amplitude spectra.}
    \label{fig:oh_amp}
\end{figure}

\vskip 5pt
\noindent \textbf{Digits-DG \& Office-Home.} We randomly select 5000 images from each domain in Digits-DG and 2000 images from each domain in Office-Home. The frequency statistics and t-SNE \cite{van2008visualizing} visualization are presented in Figs. \ref{fig:digits_amp} and \ref{fig:oh_amp}. 

\subsection{Comparison on CIFAR-10(100)(-C)}

\begin{figure}[!t]
    \centering
    \subfloat[$F_{c}$.]{
        \includegraphics[width=0.47\linewidth]{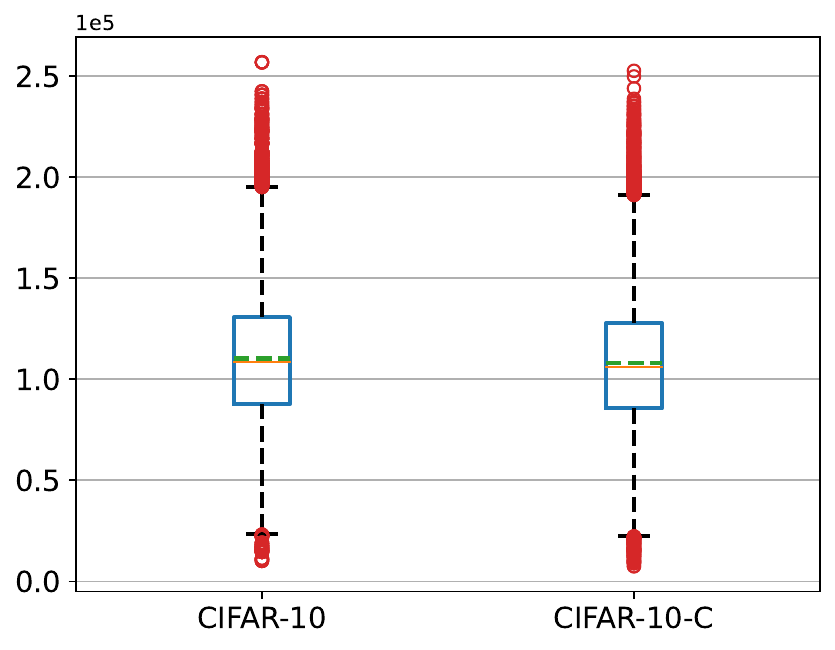}
        \label{fig:sta_cifar10(c)_fc}}
    \hfill
    \subfloat[$F_{std}$.]{
        \includegraphics[width=0.455\linewidth]{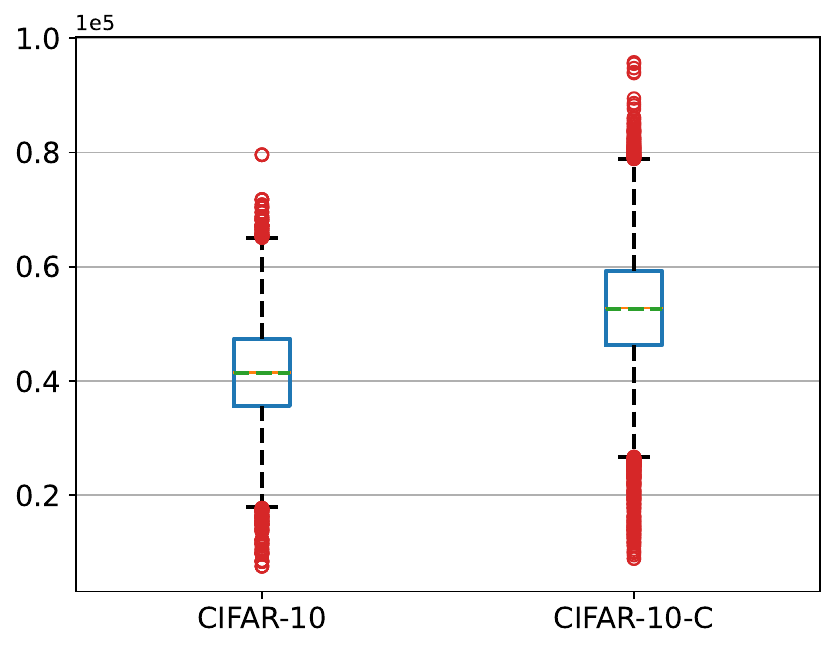}
        \label{fig:sta_cifar10(c)_fstd}}

    \subfloat[$F_{c}$.]{
        \includegraphics[width=0.47\linewidth]{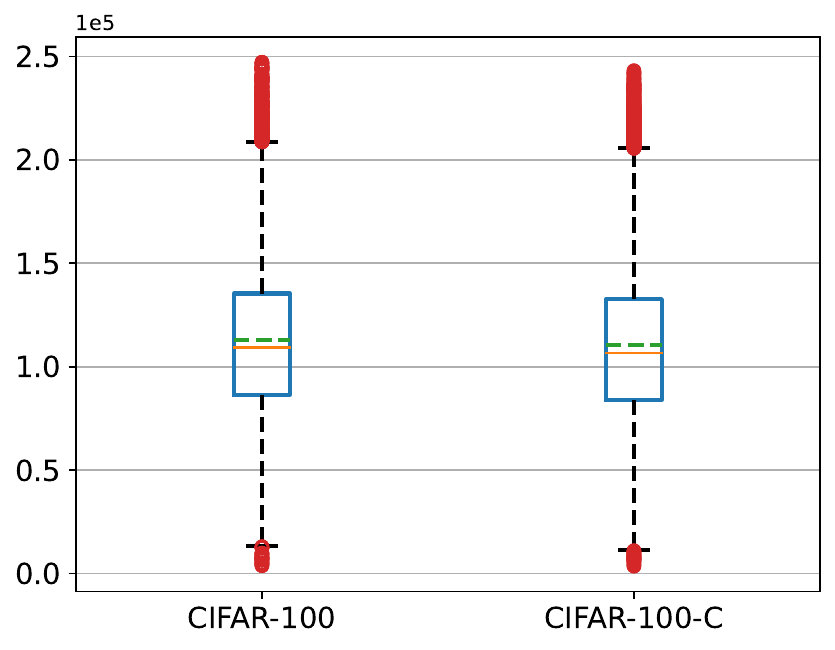}
        \label{fig:sta_cifar100(c)_fc}}
    \hfill
    \subfloat[$F_{std}$.]{
        \includegraphics[width=0.455\linewidth]{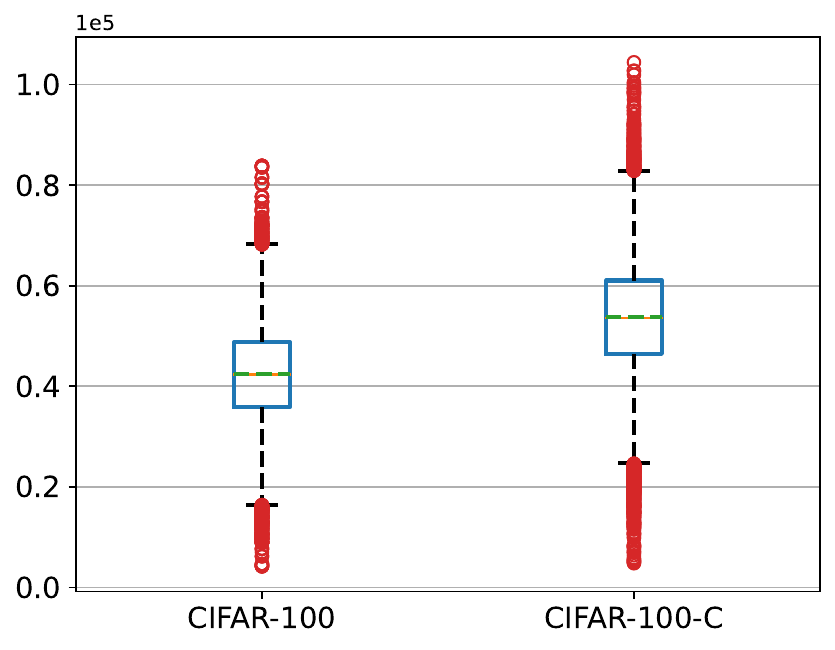}
        \label{fig:sta_cifar100(c)_fstd}}
    \caption{Boxplots of the centroid frequency and frequency standard deviation of the amplitude spectra on CIFAR(-C).}
    \label{fig:sta_cifar_amp}
\end{figure}

We randomly select 50000 images from the test split of CIFAR-10(100) (15 copies) and CIFAR-10(100)-C, respectively. The presented statistics in Fig. \ref{fig:sta_cifar_amp} further demonstrate the amplitude spectrum's sensitivity to domain shifts.

\begin{figure}[!t]
    \centering
    \subfloat[CIFAR-10(-C).]{
        \includegraphics[width=0.47\linewidth]{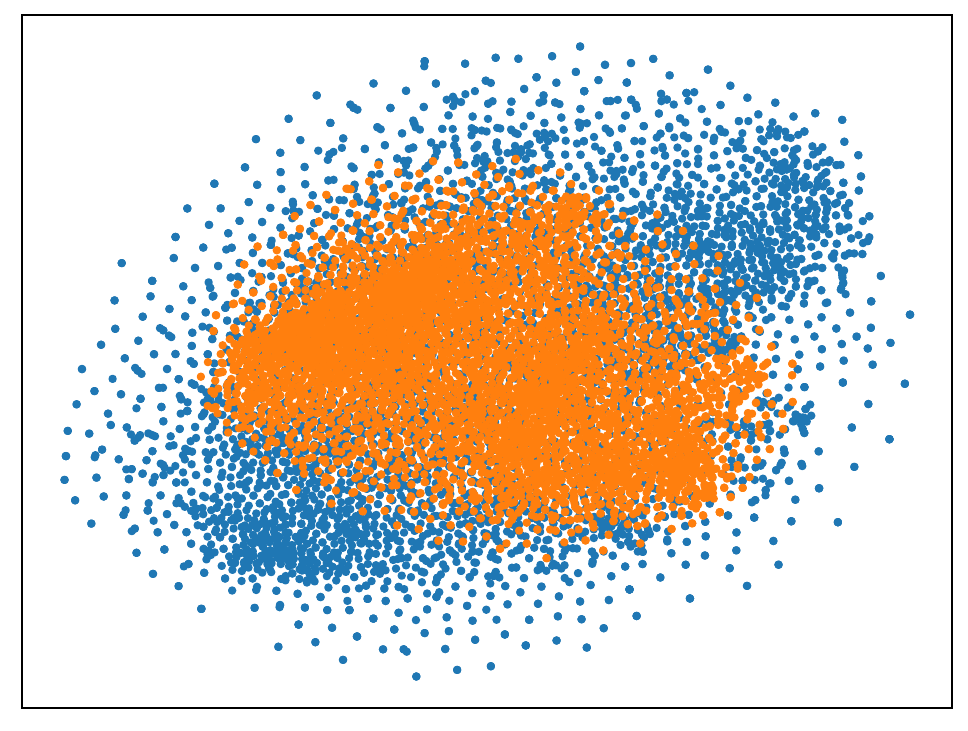}
        \label{fig:vis_cifar10}}
    \hfill
    \subfloat[CIFAR-100(-C).]{
        \includegraphics[width=0.455\linewidth]{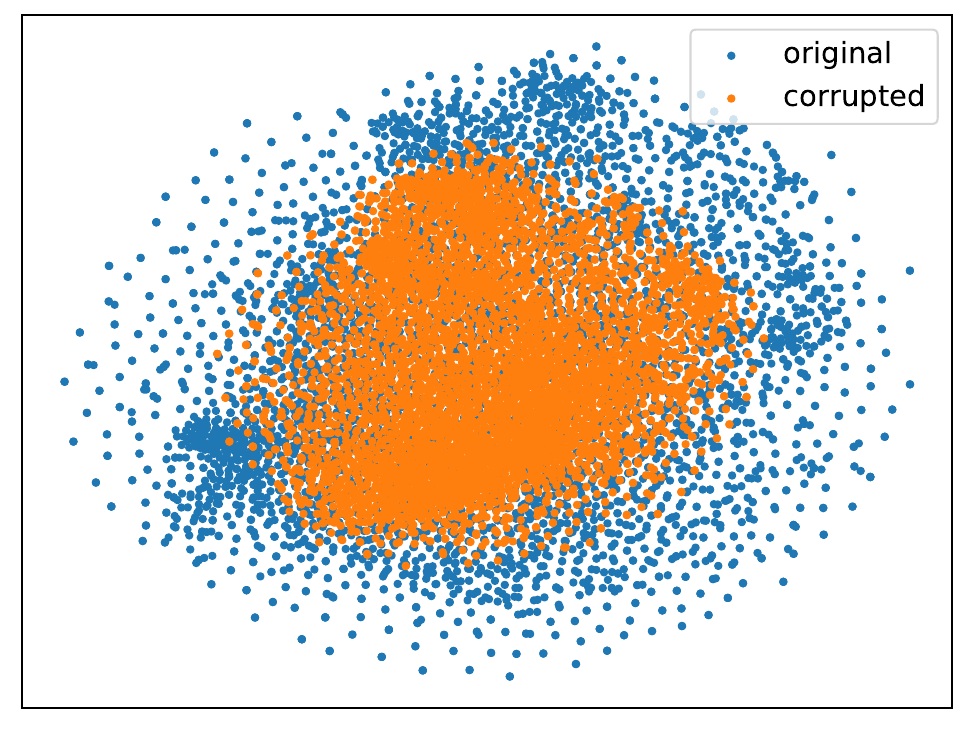}
        \label{fig:vis_cifar100}}
    \caption{t-SNE visualization of CIFAR-10(100) (blue) and CIFAR-10(100)-C (orange).}
    \label{fig:vis_cifar}
\end{figure}

Additionally, we provide the t-SNE visualization of the amplitude spectra on CIFAR(-C) in Fig. \ref{fig:vis_cifar}. The corrupted samples exhibit a notably distinct t-SNE distribution compared to the original samples.

\section{Discussion and Conclusion}
\label{sec:dis&clu}

\subsection{A Frequency Causal View for DG}
\label{sec:dg_fcv}

\begin{figure}[!t]
    \centering
    \includegraphics[width=0.95\linewidth]{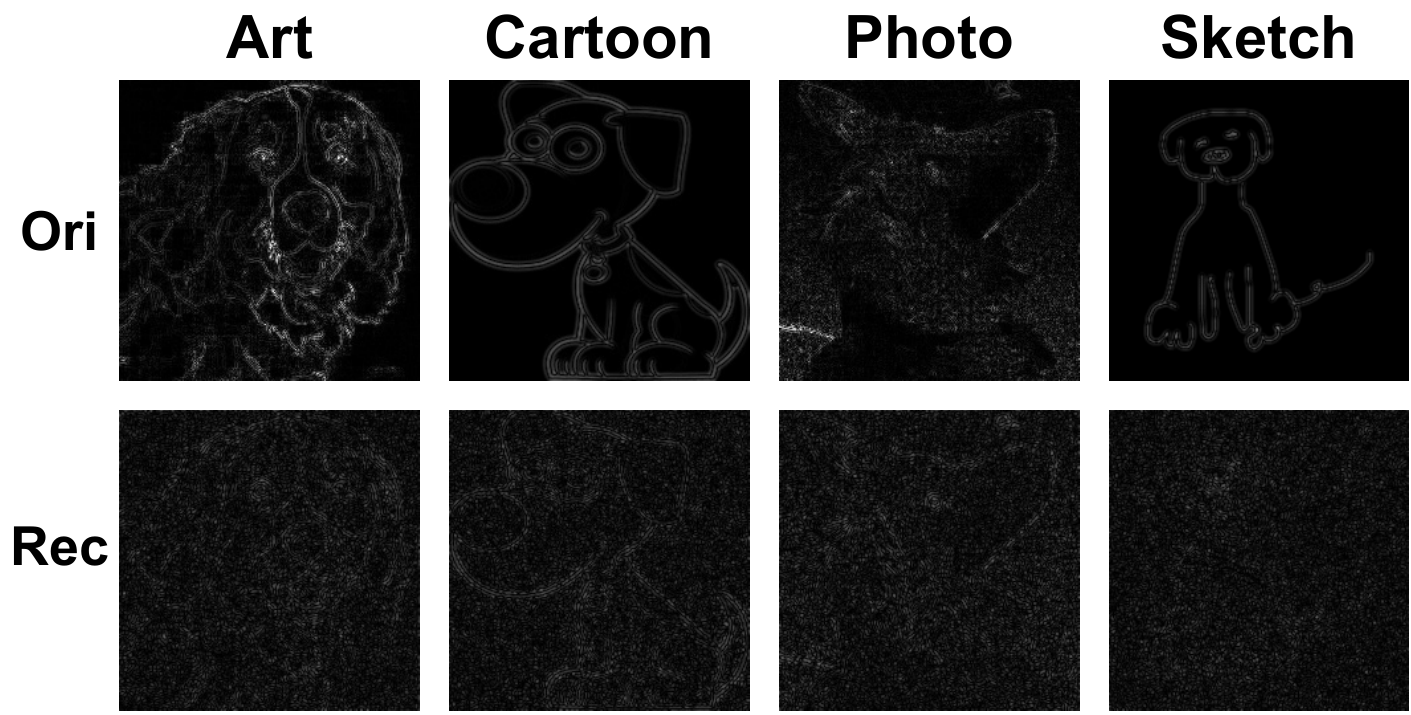}
    \caption{Edge detection results with Laplacian Operator (LO). We conduct edge detection for original and reconstructed images in the empirical experiments in Sec. \ref{sec:intro} with LO.}
    \label{fig:edge_det}
\end{figure}

As is discussed in Secs. \ref{sec:intro} and \ref{sec:dis_aly}, the amplitude spectrum is sensitive to domain shifts. For the phase spectrum, we present the cosine similarity of the phase-only reconstructed images and two edge detectors (Laplacian Operator \& Sobel Operator) in Tab. \ref{tab:edge}. Results in Tab. \ref{tab:edge} and Fig. \ref{fig:edge_det} further confirm that the phase spectrum preserves visual structures such as edges and contours, which mainly contribute to the recognition and location of objects in an image. Hence, we further explore the causal relationships between the Fourier components and DG.

\begin{table}[!t]
    \centering
    \caption{The cosine similarities between phase-only reconstructed images (Pha) and the results from the original images with Laplacian Operator (LO) and Sobel Operator (SO).}
    \resizebox{0.7\linewidth}{!}{
    \begin{tabular}{c|ccc}
	\toprule
        &  Pha\&LO&  Pha\&SO&  LO\&SO \\
	\midrule
	Cos-Sim&  0.89&  0.81&  0.87 \\
	\bottomrule
    \end{tabular}}
    \label{tab:edge}
\end{table}

\begin{figure}[!t]
    \centering
    \subfloat[General form.]{
	\label{fig:scm_g}
	\includegraphics[width=0.48\linewidth]{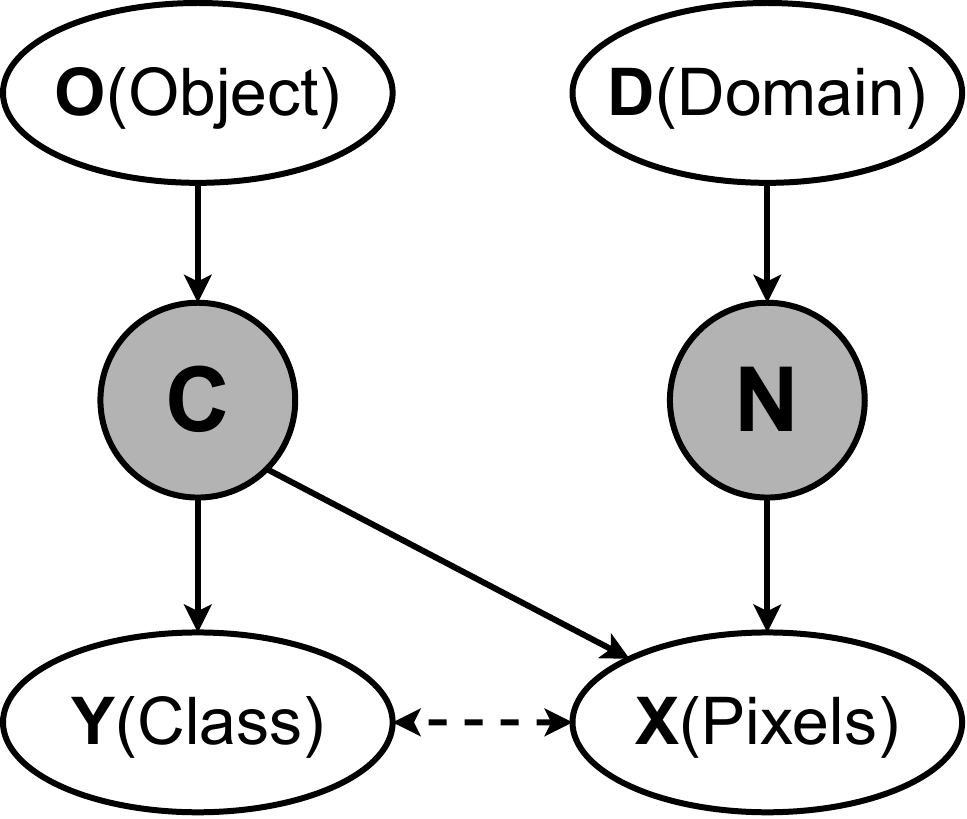}
	}
    \subfloat[Specified form.]{
	\label{fig:scm_s}
	\includegraphics[width=0.48\linewidth]{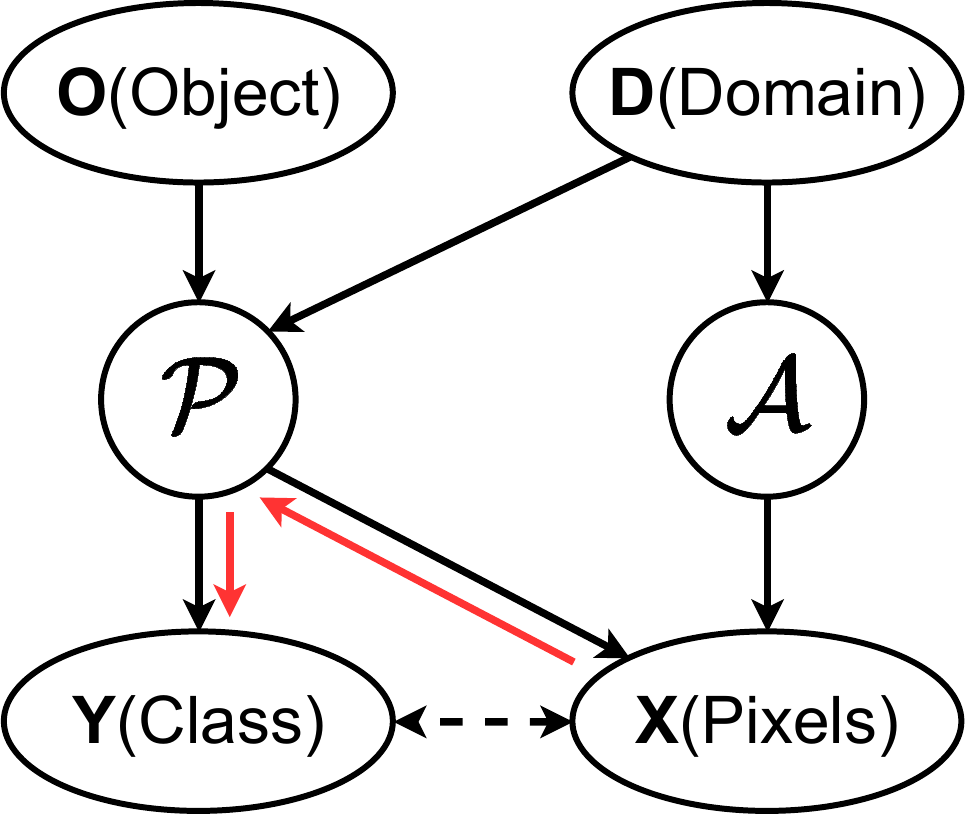}
	}
    \caption{Structural causal models (SCMs) for DG. The solid arrow denotes the parent node causes the child node; the dash arrow denotes correlation; shaded variables are unobserved.}
    \label{fig:scm} 
\end{figure}

We begin by considering DG as a domain-specific image generation task from a causal perspective. As depicted in Fig. \ref{fig:scm_g}, given information about an \textit{object} (O) and a \textit{domain} (D), the pixels of the image $X$ are constructed with both the latent embeddings of \textit{object} and \textit{domain}, whereas the category $Y$ is solely influenced by \textit{object}. In this context, we consider the latent embeddings caused by \textit{object} and \textit{domain} as \textit{causal factors} (C) and \textit{non-causal factors} (N), respectively.

Based on \textit{Reichenbach’s Common Cause Principle} \cite{reichenbach1956direction}, the Structural Causal Model (SCM) of the domain-specific image generation process can be formulated as follows:
\begin{equation}
	\begin{aligned}
		C &= U_{O}, N = U_{D},\\
		X &= g_{x}(C, N; \theta) + U_{X}, \\
		Y &= g_{y}(C; \theta) + U_{Y},
		\label{eq:scm_g}
	\end{aligned}
\end{equation}
where $U=\{U_{O}, U_{D}, U_{X}, U_{Y} \}$ denotes the \textit{exogenous variables}, and $V=\{X, Y, C, N \}$ denotes the \textit{endogenous variables}. Note that $C$ and $N$ satisfy the following conditions: \textbf{(1)} $C \not \! \perp \!\!\! \perp O$, $N \not \! \perp \!\!\! \perp D$; \textbf{(2)} $C \perp \!\!\! \perp D \mid O$, $N \perp \!\!\! \perp O \mid D$. The latter condition ensures that $C$ is invariant with the same object across different domains, and $N$ is independent of the object. Unfortunately, due to the un-observation of causal/non-causal factors, we cannot directly formulate $X = g_{x}(C, N; \theta) + U_{X}$, which remains a challenging problem for causal inference \cite{gelman2011causality} and poses a further obstacle to modeling the distribution from $X$ to $Y$. 

With reference to \cite{chen2021amplitude, mahajan2021domain} and Corollary \ref{cor:fc&ds}, we make the following assumption for components of the Fourier transform:

\begin{assumption}
    The phase component of the Fourier spectrum is dependent on both the object and domain information. The amplitude component is only dependent on the domain information.
    \label{ass:ap_csv}
\end{assumption}

By introducing the Fourier transform into causal inference in Assumption \ref{ass:ap_csv}, we can now have a formal statement for the generation process:
\begin{corollary} 
    The category of the generated image is only dependent on the phase spectrum, the pixels of the image are constructed from both the phase and amplitude spectrum.
    \label{cor:ap_real_rel}
\end{corollary}

The proof is omitted because Corollary \ref{cor:ap_real_rel} has been empirically verified in Sec. \ref{sec:intro} and previous work \cite{chen2021amplitude}. Therefore, the corollary can serve as the causal explanation for the generalizability of CNN in DG. 

With Corollary \ref{cor:ap_real_rel}, we treat the phase spectrum $\mathcal{P}$ as the \textit{semi-causal factors} (note that a causal relation between domain (D) and the phase spectrum ($\mathcal{P}$) is introduced) and the amplitude spectrum $\mathcal{A}$ as the redundant \textit{non-causal factors}. We transform the general SCM (Eq. \eqref{eq:scm_g}) into the following specified form (depicted in Fig. \ref{fig:scm_s}):
\begin{equation}
	\begin{aligned}
		\mathcal{P} &= U_{O} + U_{D}, \mathcal{A} = U_{D},\\
		X &= g_{x}(\mathcal{P}, \mathcal{A}; \theta) + U_{X}, \\
		Y &= g_{y}(\mathcal{P}; \theta) + U_{Y}.
	\end{aligned}
	\label{eq:scm_s}
\end{equation}

Therefore, the \textbf{\textit{objective}} of DG is to learn the mapping of $Y$ as $g(X_{\mathcal{P}}; \theta)$, where $g: \mathcal{X} \to \mathbb{R}^{N}$ and $X_{\mathcal{P}}$ represent the phase spectrum of input signal $X$. This process is illustrated by the red arrows in Fig. \ref{fig:scm_s}.

\subsection{Limitation}

Although our method alleviates the cross-domain impact of amplitude information, it doesn't actually build relationships between cross-domain samples, \textit{i.e.}, the intrinsic domain-invariant representation learning is weakly reflected. Besides, the spatial relationship of the embedded patches is built with an unsupervised regularization, which can further be extended with extra metrics. These limitations might result in the slight improvement of PhaMa on many DG benchmarks. Considering the great property of the Fourier transform, it is worth thinking about whether the frequency components can be embedded in the latent space for invariant representation extraction like the \textit{time-frequency encoder} in \cite{he2023domain}. Beyond domain generalization, considering the great property of the Fourier transform, it is worth thinking about whether the Fourier transform and the spatial relationships of the phase spectrum can be extended to unsupervised visual representation learning. 

\subsection{Conclusion}
In this paper, we conduct empirical experiments and distribution analysis for the Fourier components, which reveal the relationship between the Fourier spectrum and Domain Generalization (DG). We thus consider DG from a frequency perspective and present PhaMa. The main idea is to build adversarial samples and establish spatial relationships for the phase spectrum with contrastive learning. Our method shows promising results on many DG benchmarks. Moreover, we explore the causal relationships between the Fourier components and DG. We hope our work can bring more inspiration into the community.

\section{Acknowledgments}
This work was supported in part by the China Postdoctoral Science Foundation under Grant 2023M733427, Grant 2023TQ0349; in part by the Natural Science of Jiangsu Province-Basic Research Program under Grant SBK2024045473; and the Jiangsu Funding Program for Excellent Postdoctoral Talent.






\bibliographystyle{IEEEtran}
\bibliography{reference}

\end{document}